\documentclass[journal]{IEEEtran}
\usepackage[american]{babel}
\usepackage{microtype}
\usepackage[pdftex]{graphicx}
\usepackage{subfigure}
\usepackage{amsthm}

\usepackage{amsmath, amsfonts}

\usepackage{algorithm}
\usepackage{algorithmic}
\usepackage{enumerate}
\usepackage{flushend}
\usepackage{booktabs}
\usepackage{longtable}
\usepackage{setspace}
\usepackage{verbatim}
\usepackage{url}
\usepackage{cite}
\usepackage{color}

\usepackage{ulem}
\usepackage{float}
\usepackage{amssymb}
\usepackage{makecell}
\usepackage{multirow}

\begin{document}
\title{Differentiable Neural Architecture Learning for Efficient Neural Network Design}

\author{
\IEEEauthorblockN{Qingbei~Guo\IEEEauthorrefmark{1} \IEEEauthorrefmark{2},
                  Xiao-Jun~Wu\IEEEauthorrefmark{1},
                  Josef Kittler\IEEEauthorrefmark{3},~\IEEEmembership{life Member,~IEEE},
                  Zhiquan~Feng\IEEEauthorrefmark{2}}

\IEEEauthorblockA{\IEEEauthorrefmark{1}Jiangsu Provincial Engineering Laboratory of Pattern Recognition and Computational Intelligence, Jiangnan University, Wuxi 214122, China\\
\IEEEauthorrefmark{1}Corresponding author: \{wu\_xiaojun@jiangnan.edu.cn\}}

\IEEEauthorblockA{\IEEEauthorrefmark{2}Shandong Provincial Key Laboratory of Network based Intelligent Computing, University of Jinan, Jinan 250022, China}

\IEEEauthorblockA{\IEEEauthorrefmark{3}Centre for Vision, Speech and Signal Processing, University of Surrey, Guildford GU2 7XH, UK}

\thanks{Manuscript received June 28, 2019}}

\markboth{Manuscript}
{Shell \MakeLowercase{\textit{et al.}}: Bare Demo of IEEEtran.cls for IEEE Journals}

\maketitle


\begin{abstract}

Automated neural network design has received ever-increasing attention with the evolution of deep convolutional neural networks (CNNs), especially involving their deployment on embedded and mobile platforms.
One of the biggest problems that neural architecture search (NAS) confronts is that a large number of candidate neural architectures are required to train,  using, for instance, reinforcement learning and evolutionary optimisation algorithms, at a vast computation cost. Even recent differentiable neural architecture search (DNAS) samples a small number of candidate neural architectures based on the probability distribution of learned architecture parameters to select the final neural architecture.
To address this computational complexity issue, we introduce a novel \emph{architecture parameterisation} based on \emph{scaled sigmoid function}, and propose a general \emph{Differentiable Neural Architecture Learning} (DNAL) method to optimize the neural architecture without the need to evaluate candidate neural networks.
Specifically, for stochastic supernets as well as conventional CNNs, we build a new channel-wise module layer with the architecture components controlled by a scaled sigmoid function. We train these neural network models from scratch. The network optimization is decoupled into the weight optimization and the architecture optimization, which avoids the interaction between the two types of parameters and alleviates the vanishing gradient problem. We address the non-convex optimization problem of neural architecture by the continuous scaled sigmoid method with convergence guarantees.
Extensive experiments demonstrate our DNAL method delivers superior performance in terms of neural architecture search cost, and adapts to conventional CNNs (e.g., VGG16 and ResNet50), lightweight CNNs (e.g., MobileNetV2) and stochastic supernets (e.g., ProxylessNAS). The optimal networks learned by DNAL surpass those produced by the state-of-the-art methods on the benchmark CIFAR-10 and ImageNet-1K dataset in accuracy, model size and computational complexity. Our source code is available at \url{https://github.com/QingbeiGuo/DNAL.git}.

\end{abstract}


\IEEEpeerreviewmaketitle

\section{Introduction}\label{sec:Introduction}
Although convolutional neural networks have made great progress in various computer vision tasks, such as image classification~\cite{KrizhevskySH12,SimonyanZ14,HeZRS16,HuangLW16}, object detection~\cite{GirshickDDM13,Girshick15,RenHGS15} and semantic segmentation~\cite{LongSD14,RonnebergerFB15,ZhouSTL18,HeGDG17}, their deployment into many embedded applications, including robotics, self-driving cars, mobile apps and surveillance cameras, is hindered by the constrains of model size, latency and energy budget. A lot of approaches have been proposed to improve the efficiency of neural networks to handle those hardware constraints. These approaches can be divided into three categories: conventional model compression~\cite{HuangW18,LiuLSHYZ17,ZhaoNZZZT19}, lightweight network design~\cite{HuangLMW18,HowardZCKWWAA17,ZhangZLS17} and automatic neural architecture search~\cite{TanCPVSHL19,YangWCSXXT19,WuDZWSWTVJ19}. Thanks to the over-parameterisation of deep neural networks, the conventional methods compress neural network models by different compression techniques, such as pruning~\cite{LinJLDL19,HeZS17}, network quantization~\cite{CourbariauxBD15,LiL16}, tensor factorization~\cite{PengTLZXP18,YuLWT17}, and knowledge distilling~\cite{HintonVD15}. The lightweight network is heuristically constructed by designing efficient modules, including group convolutions, depthwise separable convolutions, shuffle operations, etc. Recently, in order to automatically explore the large design space, the NAS methods leverage reinforcement learning~\cite{BakerGNR16,BelloZVL17,TanCPVSHL19}, evolutionary optimisation  algorithm~\cite{RealMSSSTLK17,RealAHL19,YangWCSXXT19} and gradient-based method~\cite{LiuSY18,WuDZWSWTVJ19,WanDZHTXWYXC20} for efficient neural network search, achieving the state-of-the-art recognition performance.

However, these existing methods suffer from three problems. (1) Both the heuristic compression policy and lightweight module design require domain expertise to explore the architecture space. However, the space is so large that such hand-crafted methods cannot afford the architecture search cost. Due to the limitations imposed on the search space, the resulting neural networks are usually sub-optimal. Moreover, these methods have to take the constraint of hardware resources into account. Unfortunately, the computational complexity makes it prohibitive to produce application and hardware specific models. (2) Previous NAS methods exploit reinforcement learning and evolutionary optimisation algorithms to automatically explore the discrete search space, thus achieving the state-of-the-art recognition performance. However, such methods generate a large number of candidate neural architectures, more than 20,000 candidate neural networks across 500GPUs over 4 days in~\cite{ZophVSL17}. It is time-consuming to train and evaluate them so as to guide the neural net architecture search. (3) The existing DNAS methods relax the problem to search discrete neural architectures to optimize the probability of stochastic supernets, and allow us to explore continuous search spaces by using gradient-based methods. However, some DNAS methods still require a few candidate neural architectures to identify the best candidate by sampling based on the probability distribution of learned architecture parameters~\cite{WuDZWSWTVJ19}.

To address these problems, we introduce a novel approach which converts the discrete optimisation problem into a continuous one. This is achieved by  proposing a differentiable neural architecture learning method to automatically search for the optimal neural network parameterised in terms of a continuous scaled sigmoid function. This is the first work to apply the scaled sigmoid function to facilitate the search for efficient neural networks to the best of our knowledge. Specifically, for both conventional CNNs and stochastic supernets, we build a new channel-wise module layer controlled by the scaled sigmoid function, which can be inserted into any existing neural architectures without any special design. This module relaxes the discrete space of neural architecture search by continuous architecture representation. By progressively reducing the smoothness of the scaled sigmoid function, the continuous optimization problem is gradually turned into the original architecture optimization problem. Thus, the optimal neural architecture can be learned by using gradient-based methods with few epochs, while guaranteeing the convergence. No additional candidate neural networks are produced, significantly improving the efficiency of neural architecture search.
In order to avoid the interaction between the weight optimization and the architecture optimization, the network optimization is decoupled into the weight optimization and the architecture optimization. This also alleviates the vanishing gradient problem.
After optimizing the neural architecture, we achieve its potential representation ability by finetuning.
Extensive experiments demonstrate that our DNAL method is applicable to conventional CNNs (e.g., VGG16~\cite{SimonyanZ14} and ResNet50~\cite{HeZRS16}), lightweight CNNs (e.g., MobileNetV2~\cite{SandlerHZZC18}) and stochastic supernets (e.g., ProxylessNAS~\cite{CaiZH18}), and achieves the state-of-the-art performance on the classification task on CIFAR-10~\cite{KrizhevskyH09} and ImageNet-1K~\cite{ILSVRC15} in terms of model size, FLOPs, accuracy, and more importantly, search cost.

Our contributions can be summarized as follows.
\begin{itemize}
\item
We build a new standalone  control module based on the scaled sigmoid function to enrich the  neural network module family to enable the neural architecture optimization.
\item
We relax the discrete architecture optimization problem into a continuous one and learn the optimal neural architecture by using gradient-based methods.
\item
Our DNAL method produces no candidate neural architectures but one, thus drastically improving the efficiency of neural architecture search.
\item
It is applicable to conventional CNNs, lightweight CNNs, and stochastic supernets for automated neural architecture learning.
\item
Extensive experiments confirm that our DNAL method achieves the state-of-the-art performance on various CNN architectures, including VGG16, ResNet50, MobileNetV2, and ProxylessNAS, over the task of CIFAR-10 and ImageNet-1K classification.
\end{itemize}

The rest of this paper is organized as follows:
We first investigate the related work in Section~\ref{sec:RelatedWork}. We then present the differentiable neural architecture learning method in Section~\ref{sec:Methodology}. 
Subsequently, we demonstrate that our proposed DNAL method delivers superior performance through extensive experiments on various popular network models and datasets in Section~\ref{sec:Experiments}. We present an ablation study, which enhances the understanding of DNAL in Section~\ref{sec:Ablation_Study}. Finally, we draw the paper to a conclusion in Section~\ref{sec:Conclusion}.

\section{Related Work}\label{sec:RelatedWork}
In this section, we review various methods to yield efficient neural networks from three different perspectives.

\textbf{Conventional Model Compression.}
The conventional compression method includes pruning~\cite{LinJLDL19,HeZS17}, network quantization~\cite{CourbariauxBD15,LiL16}, tensor factorization~\cite{PengTLZXP18,YuLWT17} and knowledge distillation~\cite{HintonVD15}. Pruning is one of the most promising compression methods, which removes the redundant parts, including weights, channels and even layers, to compress neural networks based on heuristic rules. It is orthogonal to other methods to design more efficient neural networks. As the structured pruning is more efficient in reducing the parameters and computations, the channel-wise pruning methods have attracted more attention.

Huang et al. introduced a scaling factor to scale the output of specific structures, such as channels, groups, and residual blocks, and added sparsity regularizations on the scaling parameters to force them to zero~\cite{HuangW18}. The structures corresponding to zero scaling factors will be pruned as the dispensible parts to promote the design of compact CNNs.
Similarly, the scaling parameters in Batch Normalization (BN) are used to control the output of the corresponding channels without introducing any extra parameters~\cite{LiuLSHYZ17}.
In this work, we adopt  a similar approach and propose a module scaling  based on the scaled sigmoid function. The difference between the proposed approach and these methods is analyzed in detail in the supplementary material.

Most of these pruning methods require extra retraining to compensate the performance loss in accuracy due to pruning. This inevitably increases the computational cost, especially in the case of the iterative pruning and retraining scheme~\cite{LinJWZZTL20,LuoZZXWL18}. Unfortunately, these methods achieve only sub-optimal solutions, because it is prohibitive to explore the whole search space using their human-based heuristics.

\textbf{Lightweight Network Design.}
This kind of method aims to directly construct efficient neural networks by designing cheap but effective modules, rather than pruning redundant structures. Such modules have few parameters and low computation complexity with a reasonable representation capacity. They including group convolution~\cite{ZhangQXW17,HuangLMW18,WangKSC19,GuoWKF20}, depthwise separable convolution~\cite{HowardZCKWWAA17,SandlerHZZC18} and shuffle operation~\cite{ZhangZLS17}.

These human-designed modules explore only a very small portion of the design space, thus discovering only sub-optimal solutions.
Moreover, it is infeasible to design specific neural networks for every individual hardware platform. In practice, we usually reuse the neural networks for different devices to save the computation time, and adjust them to achieve a trade-off between accuracy and efficiency.

\textbf{Neural Architecture Search.}
Recently, the methodology of neural architecture search has made a significant  progress. The NAS methods automatically explore the search space to find the optimal neural architectures by different optimization methods, such as reinforcement learning, evolutionary optimization algorithms and gradient-based methods.

The early works automatically searched the optimal neural architecture based on reinforcement learning~\cite{BakerGNR16,BelloZVL17,TanCPVSHL19} and evolutionary optimization algorithms~\cite{RealMSSSTLK17,RealAHL19,YangWCSXXT19} in a discrete search  space. These methods generate thousands of candidate neural network architectures, and their validation set performance is treated as the reward or fitness to guide the search process.
However, it is time-consuming to train and evaluate those candidate architectures.
Various proxy techniques have been adopted to reduce the search cost, including the performance evaluation  on a small dataset, training for few epochs and searching few blocks~\cite{CaiZH18}.
However, they do not fundamentally solve the problem of search cost.

To solve the problem effectively, the idea of  differentiable neural architecture search was proposed in~\cite{LiuSY18,WuDZWSWTVJ19,WanDZHTXWYXC20} to optimize the network architecture by gradient-based methods.
These DNAS methods utilize the softmax function over parallel operation blocks to convert the discrete space into a continuous space, and formulate the neural architecture learning in a differentiable manner. The optimal neural architecture problem is then solved based on gradient search methods, which avoids enumerating individual network architectures and training/evaluating them separately. Nevertheless, some DNAS methods still sample multiple candidate architectures based on the learned distribution of architecture parameters, thus resulting in extra search costs~\cite{WuDZWSWTVJ19}.
We will discuss the difference between our DNAL method and the existing DNAS methods in the supplementary material.

\section{Methodology}\label{sec:Methodology}

\subsection{Problem Definition}\label{sec:Problem_Definition}

A neural network can be parameterized with two kinds of parameters, i.e., the architecture parameters, which represent the neural architecture, and the weights to generate the feature maps. Then, the problem of learning efficient neural architectures can be formulated as follows,
\begin{equation}
\label{eqn:1}
\begin{aligned}
   \underset{\mathbf{s}\in \mathcal{R}^n}{min}~\underset{a(\mathbf{s})\in \mathcal{A}}{min}~\underset{\mathbf{w}_a}{min}~{\mathcal{L}(a(\mathbf{s}), \mathbf{w}_a)}
\end{aligned}
\end{equation}

\noindent
Here, $\mathcal{R}^n$ is a high-dimensional real space, whose dimension is related to the architecture of each neural network. $\mathcal{A}$ is a discrete space of the architecture parameters $\mathbf{s}$. We aim to find an optimal architecture $a(\mathbf{s})\in \mathcal{A}$ by optimizing $\mathbf{s}$. The neural network with the optimal architecture $a(\mathbf{s})$ is trained to achieve the minimal loss $\mathcal{L}(a(\mathbf{s}), \mathbf{w}_a)$ by optimizing the weights $\mathbf{w}_a$.

\subsection{The Search Space}\label{sec:The_Search_Space}

In this paper, we build a channel-wise search space, which includes variants of conventional CNN, lightweight CNN and stochastic supernet as instances. At each layer, the conventional and lightweight CNNs define a single operation, such as conv, pooling, and etc. By choosing different channels we are able to configure different architectures. The stochastic supernet have multiple parallel blocks performing different operations at each layer, providing a greater flexibility in the choice of architecture, but creating a larger search space. The existing DNAS methods choose only one different block from multiple candidate blocks at each layer to construct a layer-wise search space~\cite{LiuSY18,WuDZWSWTVJ19,WanDZHTXWYXC20}. Unlike such DNAS methods, our approach allows to choose one or more blocks with different channels for each layer. It diversifies the structure of neural networks, which helps to improve their representation capacity~\cite{ZhangLCL17}.

The conventional CNN can be viewed as a special stochastic supernet which has a single operation block at each layer. Taking a stochastic supernet as an example, suppose that an $L$-layer stochastic supernet $\mathbf{N}$ contains $M^l$ parallel operation blocks at the $l$-th layer and each block has $N^l$ channels. The state of each channel is a binary sample space, i.e., \{0, 1\}. The zero value means the corresponding channel does not contribute to the process of inference and vice versa. The architecture search space will  include $2^{\sum_{l=1}^L{M^lN^l}}$ possible architectures. For instance for the case of  VGG16, as the total number of its channels is 4224, it contains $2^{4224} \approx 10^{1272}$ possible architectures. For deeper ResNet50, the number of possible architectures exceeds $10^{7992}$.
The search space is so large that it is not feasible to enumerate them, let alone search  for the optimal neural architecture.
Moreover, it is a challenging to solve the non-convex optimization problem to find the optimal architecture in such a  large search space.

\subsection{Differentiable Neural Architecture Learning by Continuation}\label{sec:Differentiable_Neural_Architecture_Learning_by_Continuation}

\noindent
\textbf{Scaled Sigmoid Function.}
We first introduce channel-wise architecture parameter vector $\mathbf{s}$, which serves as an indictor to represent the architecture of a neural network. Here, $\mathbf{s} = [\mathbf{s}^1, ..., \mathbf{s}^L]$, and for the $l$-th layer, $\mathbf{s}^l = [s^l_{11},...,s^l_{1N^l},...,s^l_{M^lN^l}]$.
DNAL learns an efficient neural architecture by converting the architecture parameters $\mathbf{s}$ to a binary vector $\mathbf{b}$. The binarization process can be implemented by taking the binary function $b = binary(s)$ as activation functions,
\begin{equation}
\label{eqn:2}
\begin{aligned}
   b = binary(s) = \begin{cases}
            1, & if~s \textgreater 0\\
            0, & otherwise,
   \end{cases}
\end{aligned}
\end{equation}

\noindent
where $s\in \mathbf{s}$ and $b\in \mathbf{b}$.

However, it is infeasible to train the deep neural network with the standard back-propagation (e.g., SGD), as the binary function is non-smooth and non-convex. The binary function is ill-posed at zero, which is non-differentiable, and its gradient is zero for all nonzero values, which causes the vanishing gradient problem in the neural network optimization.

\begin{figure}[t]
  \centering
  \includegraphics[trim=0mm 5mm 0mm 5mm, width=3.0in]{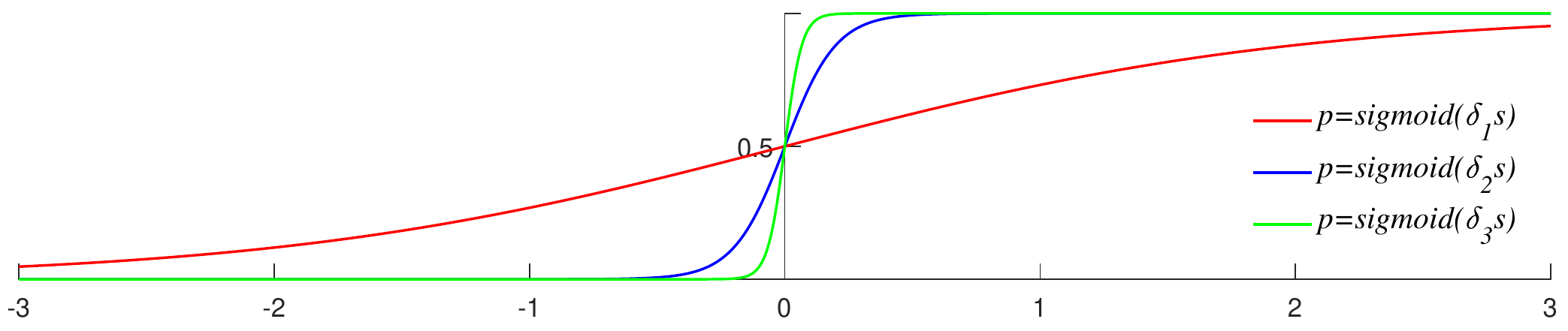}
  \caption{Scaled $sigmoid(\delta s)$ function. Green, blue and red curves show the function $p = sigmoid(\delta s)$ with hyper-parameter $\delta_1 \textless \delta_2 \textless \delta_3$. The key property is $\lim\nolimits_{\delta\to\infty}sigmoid(\delta s)=binary(s)$.}
  \label{fig:1}
\end{figure}

It is an open problem to optimize neural networks with non-smooth activation functions.
Motivated by the continuation methods~\cite{AllgowerG12}, we convert the optimization problem with ill-posed gradients to a manageable problem by smoothing the binary activation function. We find that there is a relationship between the binary function and the scaled sigmoid function which becomes binary when scale factor $\delta$ tends to infinity, as follows,
\begin{equation}
\label{eqn:3}
\begin{aligned}
   \lim\limits_{\delta\to+\infty}sigmoid(\delta s) = binary(s),
\end{aligned}
\end{equation}

\noindent
where $p = sigmoid(\delta s) = 1/(1+e^{-\delta s})$ is the scaled sigmoid function with hyper-parameter $\delta$ to control its transition from zero to one, as shown in Fig.~\ref{fig:1}. If $\delta = 1$, then it is the standard sigmoid function, which is smooth. The transition region  becomes sharper as the scale factor   $\delta$ increases. As $\delta$ approaches +$\infty$, the function is transformed into the original non-smoothed binary function.

Thanks to the key property of the scaled sigmoid function, we relax the problem of optimizing the neural architecture to the problem of optimizing the architecture parameters $\mathbf{s}$ by progressively sharpening the  scaled sigmoid function transition region.
Specifically, we begin to optimize the neural architecture with the smoothed sigmoid activation function, where $\delta_0 = 1$. By progressively increasing the scale factor $\delta$, the neural architecture will gradually converge to a solution corresponding to an optimal architecture defined by the resulting binary function. In this paper, $max(\delta) = 10^4$, which is sufficient to guarantee the convergence.  Therefore, our DNAL method can optimize the neural architecture by using gradient-based methods, while producing no additional candidate architectures.

\begin{figure*}[t]
\begin{minipage}{1\textwidth}
  \centering
  \includegraphics[trim=0mm 5mm 0mm 5mm, width=6.0in]{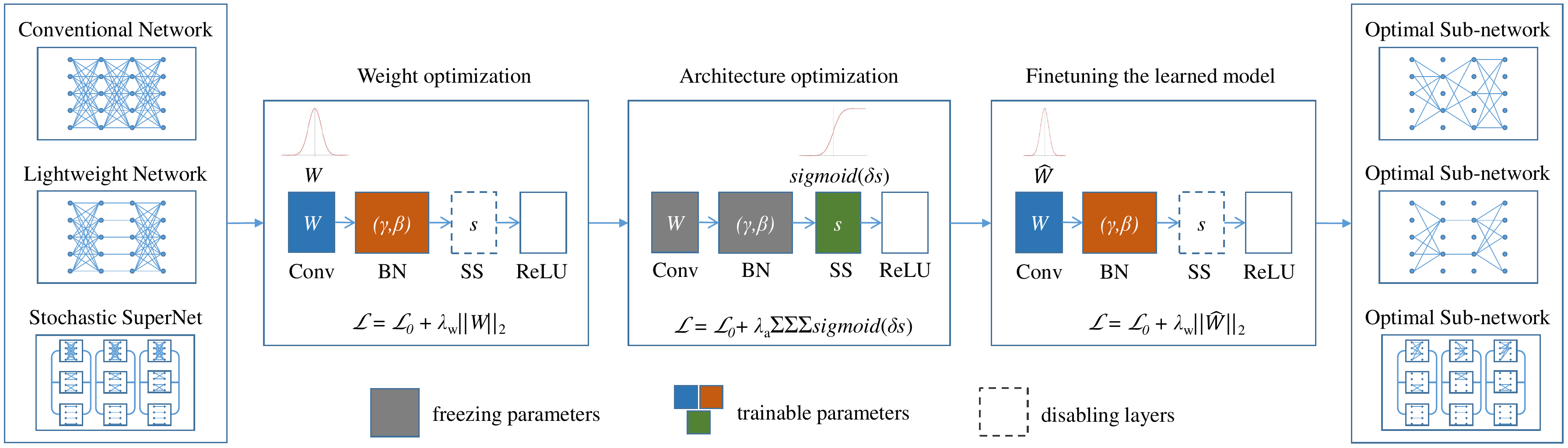}
  \caption{Differentiable neural architecture learning for efficient convolutional network design.}
  \label{fig:2}
\end{minipage}
\end{figure*}

\noindent
\textbf{Architecture Optimization.}
We build a new channel-wise module layer to incorporate the Scaled Sigmoid activation function, named by $\mathbf{SS}$, and add the $\mathbf{SS}$ transformation after the batch normalization layer, as shown in Fig.~\ref{fig:2}. The order of $\mathbf{SS}$ layer is analyzed in the ablation study presented later in the paper.

To achieve an efficient neural architecture, we define the following loss function,

\begin{equation}
\label{eqn:4}
\begin{aligned}
    \mathcal{L} = \mathcal{L}_0 + \lambda_a \sum_{l=1}^L \sum_{i=1}^{M^l} \sum_{j=1}^{N^l} {sigmoid(\delta s^l_{ij})}.
\end{aligned}
\end{equation}

\noindent
Here, the first term is the cross-entropy loss function. The second term drives the scaled sigmoid activation of each channel to zero, which tends to remove the less important channels. The hyper-parameter $\lambda_a$ is a coefficient to achieve an appropriate balance between accuracy-efficiency.

In the forward step, we calculate the output $\mathbf{\hat{x}}^l$ of the $l$-th layer as follows,

\begin{equation}
\label{eqn:5}
\begin{aligned}
    \mathbf{\hat{x}}^l = \sum_{i=1}^{M^l} ReLU(
    \left(
    \begin{array}{c}
          x^l_{i1} \\
          \vdots \\
          x^l_{iN^l}
    \end{array}
    \right)
    \odot
    \left(
    \begin{array}{c}
          sigmoid(\delta s^l_{i1}) \\
          \vdots \\
          sigmoid(\delta s^l_{iN^l})
    \end{array}
    \right)),
\end{aligned}
\end{equation}

\noindent
where $\odot$ is the symbol of Hadamard product, and $x^l_{ij}$ is the original output of the $j$-th channel of the $i$-th block in the $l$-th layer. The sigmoid function serves as a weight coefficient to scale the output of the corresponding channel. The outputs of the blocks at each layer are  aggregated as the layer output. The scale factor $\delta$ exponentially increases as the training process progresses.

In the backward step, the gradient w.r.t. the architecture parameters can be calculated as follows,

\begin{equation}
\label{eqn:6}
\begin{aligned}
    \frac{\partial\mathcal{L}}{\partial s^l_{ij}} = \frac{\partial\mathcal{L}}{\partial p^l_{ij}} \frac{\partial p^l_{ij}}{\partial s^l_{ij}} = \frac{\partial\mathcal{L}}{\partial p^l_{ij}} {\delta^l}^2 s^l_{ij} (1 - \delta^l s^l_{ij}).
\end{aligned}
\end{equation}

Our approach has a number of significant advantages. We update the architecture parameters $\mathbf{s}$ by directly using a gradient descent method. Due to the relative small number of architecture parameters, compared to the number of weights, our DNAL method exhibits a fast convergence for the architecture optimization.
Moreover, as we optimize the neural architecture directly on the target tasks, we require no proxy tasks, such as a smaller dataset for training. This ensures that the neural architecture obtained by DNAL is optimal on the target tasks.
In this work, the architecture optimization consumes about one-tenth of the resource needed for the parameter optimization, specifically 20 epochs for CIFAR-10 and 10 epochs for ImageNet, as described in Section \ref{sec:Experiments} in detail.

Once the DNAL optimization process has converged, if $sigmoid(\delta s^l_{ij}) = 1$, then the corresponding channel is significantly retained. On the contrary, if $sigmoid(\delta s^l_{ij}) = 0$, the corresponding channel is decisively pruned out as its output has no contribution to the subsequent computation. However, as this pruning process can result in  channel dimension incompatibility, we follow the technique in~\cite{WanDZHTXWYXC20}, and zero-pad the missing channels for the sake of channel dimension alignment. Moreover, after pruning, we disable all the $\mathbf{SS}$ layer, so no special operations or structures are introduced. Finally, we finetune the optimal sub-network to restore its representative ability.

\noindent
\textbf{Sequential Optimization Strategy.}
In the approach described so far, there are two challenging problems. With the increasing value  of the hyper-parameter $\delta$, the scaled sigmoid function has larger and larger saturation zone where the gradient is zero and this leads to the vanishing gradient problem. As a result, back-propagation becomes inefficient, and this is harmful to the optimization process.
The other issues is the interaction between the parameter optimization and the architecture optimization. A channel is simultaneously affected by both its architecture parameter and weights. For example, a channel has a near-zero value of scaled sigmoid function but large weights. It is arbitrary to take it as a less important channel, because the channel still may contribute considerably  to the next layer.

To tackle these two issues, we decouple the network optimization into the weight optimization and the architecture optimization. We found empirically that if the architectures are optimized from the start without a suitable initialisation, the  architecture search will tend to fall into bad local optima. Therefore, the weights are optimized first. At the weight optimization stage, we disable the $\mathbf{SS}$ layer, which means that the $\mathbf{SS}$ layer does not change the original channel output, i.e., $sigmoid(\delta s)$ = 1 for each channel, and we use SGD to learn only the weights. When optimizing the neural architecture, we freeze the trainable layers, including convolutional layers and BN layers, and focus only on the architecture parameters also by SGD. Using the sequential optimization strategy alleviates the vanishing gradient problem, as well as reducing the interaction weights and architecture parameters. We will compare the sequential optimization strategy with the joint optimization in the following ablation study.

\noindent
\textbf{DNAL Algorithm.}
We depict the optimization process of our DNAL method in Alg.~\ref{alg:1}. The whole process consists of three stages: (1) Weight optimization stage. We learn only the weights by SGD with weight decay, while disabling the $\mathbf{SS}$ layer. (2) Architecture optimization stage. When learning architecture parameters, the weights are frozen. The scaled sigmoid technique is used to optimize the architecture parameters by SGD. (3) Finetuning stage. After pruning the channels with $sigmoid(\delta s)$ = 0, we finetune the derived neural network to achieve a better accuracy.

\begin{algorithm}[htbp]
  \small
  \begin{algorithmic}[1]
    \REQUIRE ~~\\
        The stochastic supernet $\mathbf{N}$ with $\mathbf{SS}$ layers, a sequence $1=\delta_1 \textless \cdots \textless \delta_n = +\infty$.
    \ENSURE ~~\\
        the efficient neural network $\mathbf{\hat{N}}$.
    \STATE randomly initializing the weights $W$, and disabling the $\mathbf{SS}$ layers
    \FOR {each epoch $i$ = 1 to $m$}
        \STATE optimizing $W$ by SGD with respect to $\mathcal{L}$ = $\mathcal{L}_0$ + $\lambda_w \left\|W\right\|_2$
    \ENDFOR
    \STATE enabling the $\mathbf{SS}$ layers, initializing $s$ = 0, and freezing the weights $W$
    \FOR {each epoch $i$ = 1 to $n$}
        \STATE optimizing $S$ by SGD with respect to $\mathcal{L}$ = $\mathcal{L}_0$ + $\lambda_a \sum\sum\sum sigmoid(\delta_i s)$
    \ENDFOR
    \STATE pruning the channels with $sigmoid(\delta s)$ = 0
    \STATE disabling the $\mathbf{SS}$ layers
    \STATE finetuning the searched network $\mathbf{\hat{N}}$ by SGD with respect to $\mathcal{L}$ = $\mathcal{L}_0$ + $\lambda_w \left\|\hat{W}\right\|_2$
  \end{algorithmic}
  \caption{Differentiable neural architecture learning procedure.}
  \label{alg:1}
\end{algorithm}

\section{Analysis}\label{sec:Analysis}

In this section, we investigate the differences between DNAL and other related methods by comparison.

\noindent
\textbf{Comparison with other DNAS methods.}
Although our DNAL method adopts the same gradient-based approach as the existing DNAS methods to optimize the neural architecture, there are major differences between them in following three respects.

First, the existing DNAS methods applied the softmax function to learn the relative probability of each operation block, and then retain the block with the maximum probability to construct the optimal architecture, while abandoning the other components~\cite{LiuSY18,WuDZWSWTVJ19,WanDZHTXWYXC20}. DNAL utilizes the scaled sigmoid function to learn the absolute probability of each channel. After converging to the original binary problem, we preserve the channels with probability 1, while removing the channels with probability 0.

Second, the existing DNAS methods choose the most likely operation block from multiple candidate operation blocks for each layer to construct the optimal neural architecture, which means each layer contains only one operation block~\cite{LiuSY18,WuDZWSWTVJ19,WanDZHTXWYXC20}. However, DNAL learns more general structure, where each layer may contain different operation blocks with different channels. Thus, DNAL increases the search space size by orders of magnitude. This helps to improves the accuracy of neural networks, which is experimentally confirmed in the following section.

Third, after finishing the architecture search, some DNAS methods still sample several candidate architectures based on the distribution of architecture parameters, and select the best one by training them from scratch~\cite{WuDZWSWTVJ19}. Our DNAL method yields directly the optimal architecture by the proposed method, producing no additional candidate architectures. This significantly reduces the computational cost.

%

\noindent
\textbf{Comparison with other scaling methods.}
In DNAL, we introduce the scaled sigmoid function as a mechanism to scale the output of each channel. The proposed method is significantly different from other scaling methods.

First, the existing scaling methods consider the scale factor as a coefficient to scale the output of some specific structures~\cite{HuangW18}, including channels, groups and blocks. Some methods leveraged the learnable scale factor of BN without introducing extra parameters~\cite{LiuLSHYZ17,ZhaoNZZZT19}. We define an architecture parameter $s \in \mathbb{R}$, and use its scaled sigmoid function $sigmoid(\delta s)\in [0, 1]$, as the scale factor in a probabilistic way.

Second, these scaling methods impose a sparsity constraint on the scaling parameters to push them infinitely close to zero, and then prune the structures corresponding to zero or near-zero out. However, such pruning may degrade the performance.
By contrast, our DNAL method forces the sparsity constraint on the scaled sigmoid function rather than the architecture parameters, driving them into the negative saturation zone, i.e., $sigmoid(\delta s) = 0$. Removing the channels with $sigmoid(\delta s) = 0$ causes no accuracy loss due to the consistency of the  binary expression with the pruned structure configuration.

\section{Experiments}\label{sec:Experiments}
In this section, we empirically evaluate the proposed DNAL method on CIFAR-10~\cite{KrizhevskyH09} and ImageNet-1K~\cite{ILSVRC15} for classification by using state-of-the-art CNN architectures, which include conventional CNNs (e.g., VGG~\cite{SimonyanZ14} and ResNet~\cite{HeZRS16}), lightweight CNNs (e.g., MobileNetV2~\cite{SandlerHZZC18}) and stochastic supernets (e.g., ProxylessNAS~\cite{CaiZH18}).
We use PyTorch~\cite{pytorch17} to implement the proposed DNAL method.

\subsection{Datasets}\label{sec:Datasets}
\noindent
\textbf{CIFAR-10.} CIFAR-10 is a popular dataset of tiny images with 10 classes, and contains 50,000 images for training and 10,000 images for testing.

\noindent
\textbf{ImageNet-1K.} ImageNet-1K is a subset of the ImageNet Large Scale Visual Recognition Challenge dataset. It contains 1.2M training images and 50K validation images as testing images, and is categorized into 1000 classes.

\subsection{Classification on CIFAR-10}\label{sec:Classification_on_CIFAR10}
We evaluate the recognition performance on CIFAR-10, comparing against several popular convolutional neural networks, such as VGG16, ResNet56 and MobileNetV2.

\noindent
\textbf{Implementation.}
We use a variation of VGG16, as in~\cite{LinJYZCYHD19}. In the first weight optimization stage, the initial model is trained for 100 epochs, and the learning rate is fixed to 0.1. In the architecture optimization stage, we learn the optimal neural architecture for 20 epochs with a constant learning rate 0.1. The scale factor $\delta$ grows from 1 to $10^4$ exponentially. In the last finetuning stage, the learning rate schedule is 0.1 for the first 30 epochs, 0.01 until the 80th epoch, and 0.001 to the 130the epoch. The batch-size is set to be 128, weight decay 1e-4 and momentum 0.9.
For ResNet56, we follow the same setting as VGG16, except for reducing the batch-size to 64. For MobileNetV2, the setting is different from VGG16. The weight optimization stage lasts for 150 epochs. The learned neural network is finetuned for 180 epochs. The learning rate is divided by 10 after 30 and 105 epochs. The weight decay is 4e-5.

\noindent
\textbf{VGG16.}
Tab.~\ref{tab:1} shows the performance of different compression methods. Compared with Variational-pruning, DNAL achieves better accuracy (93.53\% vs. 93.18\%) with similar compression and acceleration rates.
Compared with GAL-0.1, DNAL provides larger reductions in FLOPs and Params, while achieving better accuracy (93.75\% vs. 93.42\%).
Compared with HRank, DNAL is significantly better in all respects (61.23 vs. 73.70 in FLOPs, 0.60 vs. 1.78 in Params and 92.33\% vs. 91.23\% in top-1 accuracy).
In addition, we tried for a higher compression and acceleration rate of up to about 90$\times$ and 20$\times$, respectively, for mirconet, achieving 89.27\% top-1 accuracy and 99.51\% top-5 accuracy. It demonstrates the ability of our proposed DNAL to find a more efficient neural networks.

\begin{table}[!t]
  \scriptsize
  \caption{Comparison of DNAL applied to VGG16 with different methods on CIFAR-10.}
  \label{tab:1}
  \centering
  \setlength{\tabcolsep}{1.8mm}{
  \begin{tabular}{l|rr|rr}
    \specialrule{0.10em}{0pt}{0pt}
    Model &
    \makecell*[c]{FLOPs\\ (M)}  &
    \makecell*[c]{Params\\ (M)} &
    \makecell*[c]{Top-1\\ (\%)} &
    \makecell*[c]{Top-5\\ (\%)} \\
    \specialrule{0.10em}{0pt}{0pt}
    Baseline                               & 313.47(1.00$\times$)   & 14.99(1.00$\times$)  & 93.77           & 99.73  \\
    \specialrule{0.08em}{0pt}{0pt}
    \textbf{DNAL}($\lambda_a$=1e-5)        & 211.89(1.48$\times$)   &  5.51(2.72$\times$)  & 93.82           & 99.71  \\
    \textbf{DNAL}($\lambda_a$=5e-5)        & 195.14(1.61$\times$)   &  3.73(4.02$\times$)  & 93.53           & 99.77  \\
    Variational-pruning~\cite{ZhaoNZZZT19} &    190(1.65$\times$)   &  3.92(3.82$\times$)  & 93.18           & -      \\
    GAL-0.1~\cite{LinJYZCYHD19}            & 171.89(1.82$\times$)   &  2.67(5.61$\times$)  & 93.42           & -      \\
    \textbf{DNAL}($\lambda_a$=1e-4)        & 161.97(1.94$\times$)   &  2.10(7.14$\times$)  & 93.75           & 99.72  \\
    HRank~\cite{LinJWZZTL20}               &  73.70(4.25$\times$)   &  1.78(8.42$\times$)  & 91.23           & -      \\
    \textbf{DNAL}($\lambda_a$=2e-4)        &  61.23(5.12$\times$)   &  0.60(24.98$\times$) & 92.33           & 99.69  \\
    \textbf{DNAL}($\lambda_a$=3e-4)        &  29.77(10.53$\times$)  &  0.29(51.69$\times$) & 89.93           & 99.62  \\
    \textbf{DNAL}($\lambda_a$=4e-4)        &  22.04(14.22$\times$)  &  0.24(62.46$\times$) & 89.92           & 99.41  \\
    \textbf{DNAL}($\lambda_a$=5e-4)        &  16.65(18.83$\times$)  &  0.17(88.18$\times$) & 89.27           & 99.51  \\
    \specialrule{0.10em}{0pt}{0pt}
  \end{tabular}}
\end{table}

\noindent
\textbf{ResNet56.}
We show the results for ResNet56 in Tab.~\ref{tab:2}. The proposed DNAL outperforms HRank in all respects (83.11 vs. 88.72 in FLOPs, 0.59 vs. 0.71 in Params and 93.75\% vs. 93.52\% in top-1 accuracy).
With a similar model size and computation complexity, we achieve better accuracy than NISP (93.75\% vs. 93.01\%).
DNAL yields 1.3\% higher top-1 accuracy and about 2$\times$ faster speedup than AMC, and also yields 0.32\% and 1.62\% higher top-1 accuracy than KSE and GAL-0.8 with a smaller model size and faster speedup, respectively.
To explore more efficient neural models, we further compress the neural network, up to more than 70$\times$ for model size and more than 120$\times$ for computation complexity, achieving 83.48\% and 99.19\% in top-1 and top-5 accuracy, respectively.

\begin{table}[!t]
  \scriptsize
  \caption{Comparison of DNAL applied to ResNet56 with different methods on CIFAR-10.}
  \label{tab:2}
  \centering
  \setlength{\tabcolsep}{1.8mm}{
  \begin{tabular}{l|rr|rr}
    \specialrule{0.10em}{0pt}{0pt}
    Model &
    \makecell*[c]{FLOPs\\ (M)}  &
    \makecell*[c]{Params\\ (M)} &
    \makecell*[c]{Top-1\\ (\%)} &
    \makecell*[c]{Top-5\\ (\%)} \\
    \specialrule{0.10em}{0pt}{0pt}
    Baseline                               & 125.49(1.00$\times$)  &  0.85(1.00$\times$)   & 94.15           & 99.91  \\
    \specialrule{0.08em}{0pt}{0pt}
    \textbf{DNAL}($\lambda_a$=1e-5)        &  93.94(1.34$\times$)  &  0.66(1.29$\times$)   & 93.76           & 99.91  \\
    HRank~\cite{LinJWZZTL20}               &  88.72(1.41$\times$)  &  0.71(1.20$\times$)   & 93.52           & -      \\
    \textbf{DNAL}($\lambda_a$=5e-5)        &  83.11(1.51$\times$)  &  0.59(1.44$\times$)   & 93.75           & 99.87  \\
    NISP~\cite{YuLCLMHGLD18}               &  81.00(1.55$\times$)  &  0.49(1.73$\times$)   & 93.01           & -      \\
    AMC~\cite{HeLLWLH18}                   &   62.7(2.00$\times$)  &                  -    &  91.9           & -      \\
    KSE(G=4)~\cite{LiLZLDWHJ19}            &     60(2.09$\times$)  &  0.43(1.98$\times$)   & 93.23           & -      \\
    KSE(G=5)~\cite{LiLZLDWHJ19}            &     50(2.51$\times$)  &  0.36(2.36$\times$)   & 92.88           & -      \\
    GAL-0.8~\cite{LinJYZCYHD19}            &  49.99(2.51$\times$)  &  0.29(2.93$\times$)   & 91.58           & -      \\
    \textbf{DNAL}($\lambda_a$=1e-4)        &  36.94(3.40$\times$)  &  0.25(3.40$\times$)   & 93.20           & 99.89  \\
    HRank~\cite{LinJWZZTL20}               &  32.52(3.86$\times$)  &  0.27(3.15$\times$)   & 90.72           & -      \\
    \textbf{DNAL}($\lambda_a$=2e-4)        &   8.63(14.54$\times$) & 0.060(14.17$\times$)   & 89.31          & 99.66  \\
    \textbf{DNAL}($\lambda_a$=3e-4)        &   3.44(36.48$\times$) & 0.022(38.64$\times$)  & 85.83           & 99.45  \\
    \textbf{DNAL}($\lambda_a$=4e-4)        &   2.38(52.73$\times$) & 0.013(65.38$\times$)  & 84.07           & 99.31  \\
    \textbf{DNAL}($\lambda_a$=5e-4)        &   1.68(74.70$\times$) &0.007(121.43$\times$)  & 83.48           & 99.19  \\
    \specialrule{0.10em}{0pt}{0pt}
  \end{tabular}}
\end{table}

\noindent
\textbf{MobileNetV2.}
The results for MobileNetV2 are shown in Tab.~\ref{tab:3}. Since the network is already very computationally efficient, it is interesting to see whether it can be compressed further.
Compared with FLGC, DNAL again demonstrates its outstanding ability to find efficient neural networks. When the acceleration rate is less than 2$\times$, our DNAL achieves the best top-1 accuracy (94.30\%),  close to the baseline.
When the acceleration rate exceeds 2$\times$, our DNAL achieves 94.01\% top-1 accuracy, which is still better than FLGC.
Although it is much harder to further compress the lightweight model, DNAL still manages to obtain 87.85\% top-1 accuracy and 99.62\% top-5 accuracy with an acceleration rate of about 20$\times$ and compression rate of roughly 30$\times$.

\begin{table}[!t]
  \scriptsize
  \caption{Comparison DNAL applied to MobileNetV2 with different methods on CIFAR-10.}
  \label{tab:3}
  \centering
  \setlength{\tabcolsep}{1.8mm}{
  \begin{tabular}{l|rr|rr}
    \specialrule{0.10em}{0pt}{0pt}
    Model &
    \makecell*[c]{FLOPs\\ (M)}  &
    \makecell*[c]{Params\\ (M)} &
    \makecell*[c]{Top-1\\ (\%)} &
    \makecell*[c]{Top-5\\ (\%)} \\
    \specialrule{0.10em}{0pt}{0pt}
    Baseline                            &  91.17(1.00$\times$) &  2.30(1.00$\times$)  & 94.31           & 99.90  \\
    \specialrule{0.08em}{0pt}{0pt}
    FLGC(G=2)~\cite{WangKSC19}          &     79(1.15$\times$) &  1.18(1.95$\times$)  & 94.11           & -      \\
    FLGC(G=3)~\cite{WangKSC19}          &     61(1.49$\times$) &  0.85(2.71$\times$)  & 94.20           & -      \\
    \textbf{DNAL}($\lambda_a$=1e-5)     &  59.47(1.53$\times$) &  1.43(1.61$\times$)  & 94.17           & 99.89  \\
    \textbf{DNAL}($\lambda_a$=5e-5)     &  54.98(1.66$\times$) &  1.20(1.92$\times$)  & 94.30           & 99.86  \\
    FLGC(G=4)~\cite{WangKSC19}          &   51.5(1.77$\times$) &  0.68(3.38$\times$)  & 94.16           & -      \\
    FLGC(G=5)~\cite{WangKSC19}          &     46(1.98$\times$) &  0.58(3.97$\times$)  & 93.88           & -      \\
    FLGC(G=6)~\cite{WangKSC19}          &   42.5(2.15$\times$) &  0.51(4.51$\times$)  & 93.67           & -      \\
    FLGC(G=7)~\cite{WangKSC19}          &     40(2.28$\times$) &  0.46(5.00$\times$)  & 93.66           & -      \\
    FLGC(G=8)~\cite{WangKSC19}          &     38(2.40$\times$) &  0.43(5.35$\times$)  & 93.09           & -      \\
    \textbf{DNAL}($\lambda_a$=1e-4)     &  36.63(2.49$\times$) &  0.65(3.54$\times$)  & 94.01           & 99.89  \\
    \textbf{DNAL}($\lambda_a$=2e-4)     &  13.35(6.83$\times$) & 0.20(11.50$\times$)  & 91.96           & 99.91  \\
    \textbf{DNAL}($\lambda_a$=3e-4)     &  7.81(11.67$\times$) & 0.12(19.17$\times$)  & 90.65           & 99.82  \\
    \textbf{DNAL}($\lambda_a$=4e-4)     &  5.40(16.88$\times$) &0.096(23.96$\times$)  & 88.83           & 99.76  \\
    \textbf{DNAL}($\lambda_a$=5e-4)     &  4.50(20.26$\times$) &0.081(28.40$\times$)  & 87.85           & 99.62  \\
    \specialrule{0.10em}{0pt}{0pt}
  \end{tabular}}
\end{table}

\subsection{Classification on ImageNet-1K}\label{sec:Classification_on_ImageNet}
We further conduct the experiments for several popular CNNs, i.e., VGG16, ResNet50, MobileNetV2 and ProxylessNAS, to evaluate the recognition performance on the large-scale ImageNet-1K.

\noindent
\textbf{Implementation.}
Both VGG16 and ResNet50 are initially trained for 30 epochs with a fixed learning rate 0.1. Then, we optimize the architecture parameters for 10 epochs with a constant learning rate 0.01. The optimal neural architectures are learned by the scaled sigmoid function method with the exponential scale factor $\delta$ ranging from 1 to $10^4$. The architecture learning process is the same for other models, i.e., MobileNetV2 and ProxylessNAS. Finally, the derived neural networks are finetuned for 70 epochs. We use SGD with a mini-batch size of 256, a weight decay of 0.0001 and a momentum of 0.9. The learning rate is divided by 10 at 10 and 40 epochs.
We train MobileNetV2 from scratch for 80 epochs. The learning rate starts from 0.1, and is tuned with a cosine decaying schedule. The derived neural networks are finetuned for 90 epochs. The weight decay is set to 4e-5. The other hyper-parameters are the same for both VGG16 and ResNet50. We remove the dropout from the last classifier layer of MobileNetV2.
We use the same search space as ProxylessNAS but without the zero operation. We train the stochastic supernet for 100 epochs from scratch. The learning rate is set to 0.05 initially, and is tuned with the same decaying schedule as used for MobileNetV2. We finetune the derived neural networks for 110 epochs by SGD with a mini-batch size of 96. The other hyper-parameters are the same to MobileNetV2.

\noindent
\textbf{VGG16.}
Tab.~\ref{tab:4} shows the performance of different methods. Compared with GDP, our DNAL method achieves a faster acceleration rate (3.23$\times$ vs. 2.42$\times$) and 1\% higher top-1 accuracy (69.80\% vs. 68.80\%). Compared with both ThiNet and SSR, DNAL provides significantly better parameter reduction (77.05 vs. 131.44 and 77.05 vs. 126.7), while maintaining a comparable performance in top-1 accuracy.

\begin{table}[!t]
  \scriptsize
  \caption{Comparison of DNAL applied to  VGG16 with different methods on ImageNet.}
  \label{tab:4}
  \centering
  \setlength{\tabcolsep}{1.8mm}{
  \begin{tabular}{l|rr|rr|r}
    \specialrule{0.10em}{0pt}{0pt}
    Model &
    \makecell*[c]{FLOPs\\ (G)}  &
    \makecell*[c]{Params\\ (M)} &
    \makecell*[c]{Top-1\\ (\%)} &
    \makecell*[c]{Top-5\\ (\%)} &
    \makecell*[c]{Search Cost\\ (Epochs)} \\
    \specialrule{0.10em}{0pt}{0pt}
    Baseline                            &  15.47(1.00$\times$)  & 138.37(1.00$\times$)  & 76.13  & 92.86  & 90    \\
    \specialrule{0.08em}{0pt}{0pt}
    GDP~\cite{LinJLWHZ18}               &    7.5(2.06$\times$)  &                    -  & 69.88  & 89.16  & 90+30+20    \\
    GDP~\cite{LinJLWHZ18}               &    6.4(2.42$\times$)  &                    -  & 68.80  & 88.77  & 90+30+20    \\
    ThiNet~\cite{LuoZZXWL18}            &   4.79(3.23$\times$)  & 131.44(1.05$\times$)  & 69.74  & 89.41  & 196+48     \\
    \textbf{DNAL}($\lambda_a$=1e-4)     &   4.69(3.30$\times$)  &  77.05(1.80$\times$)  & 69.80  & 89.42  & 30+10+70\\
    SSR-L2,1~\cite{LinJLDL19}           &    4.5(3.44$\times$)  &  126.7(1.09$\times$)  & 69.80  & 89.53  & -     \\
    SSR-L2,0~\cite{LinJLDL19}           &    4.5(3.44$\times$)  &  126.2(1.10$\times$)  & 69.99  & 89.42  & -     \\
    GDP~\cite{LinJLWHZ18}               &    3.8(4.07$\times$)  &                    -  & 67.51  & 87.95  & 90+30+20     \\
    SSS~\cite{HuangW18}                 &    3.8(4.07$\times$)  &  130.5(1.06$\times$)  & 68.53  & 88.20  & 100     \\
    ThiNet~\cite{LuoZZXWL18}            &   3.46(4.47$\times$)  & 130.50(1.06$\times$)  & 69.11  & 88.86  & 196+48     \\
    \specialrule{0.10em}{0pt}{0pt}
  \end{tabular}}
\end{table}

\noindent
\textbf{ResNet50.}
For ResNet50, we summarize the performance comparison with various methods in Tab.~\ref{tab:5}. We observe that DNAL outperforms SSR by a significant margin in all respects. Similarly, DNAL achieves better performance than both GDP and GAL.  Compared with ThiNet-50, DNAL achieves 1.62\% higher top-1 accuracy with similar FLOPs and parameter reductions (1.75 vs. 1.71 in FLOPs and 12.75 vs. 12.38 in Params). Similar observations can be found when comparing with HRank. DNAL yields 0.88\% and 1.07\% higher top-1 accuracy than HRank, respectively, while maintaining lower computation complexity. Compared with ABCPruner, DNAL is a slightly better in accuracy (73.65\% vs. 73.52\%) with  fewer FLOPs (1.75 vs. 1.79).

\begin{table}[!t]
  \scriptsize
  \caption{Comparison of DNAL applied to  ResNet50 with different methods on ImageNet.}
  \label{tab:5}
  \centering
  \setlength{\tabcolsep}{1.8mm}{
  \begin{tabular}{l|rr|rr|r}
    \specialrule{0.10em}{0pt}{0pt}
    Model &
    \makecell*[c]{FLOPs\\ (G)}  &
    \makecell*[c]{Params\\ (M)} &
    \makecell*[c]{Top-1\\ (\%)} &
    \makecell*[c]{Top-5\\ (\%)} &
    \makecell*[c]{Search Cost\\ (Epochs)} \\
    \specialrule{0.10em}{0pt}{0pt}
    Baseline                           &  4.09(1.00$\times$)  & 25.55(1.00$\times$)  & 75.19  & 92.56  & 90    \\
    \specialrule{0.08em}{0.5pt}{0pt}
    \textbf{DNAL}($\lambda_a$=5e-5)    &  2.07(1.98$\times$)  & 15.34(1.67$\times$)  & 74.07  & 92.02  & 30+10+70\\
    SSR-L2,1~\cite{LinJLDL19}          &   1.9(2.15$\times$)  &  15.9(1.61$\times$)  & 72.13  & 90.57  & -  \\
    SSR-L2,0~\cite{LinJLDL19}          &   1.9(2.15$\times$)  &  15.5(1.65$\times$)  & 72.29  & 90.73  & -  \\
    GDP~\cite{LinJLWHZ18}              &  1.88(2.18$\times$)  &                   -  & 71.89  & 90.71  & 90+30+20   \\
    GAL-0.5-joint~\cite{LinJYZCYHD19}  &  1.84(2.22$\times$)  & 19.31(1.32$\times$)  & 71.80  & 90.82  & 90+60     \\
    ABCPruner~\cite{LinJZZWT20}        &  1.79(2.28$\times$)  & 11.24(2.27$\times$)  & 73.52  & 91.51  & 12+90\\
    \textbf{DNAL}($\lambda_a$=6e-5)    &  1.75(2.34$\times$)  & 12.75(2.00$\times$)  & 73.65  & 91.74  & 30+10+70\\
    ThiNet-50~\cite{LuoZZXWL18}        &  1.71(2.39$\times$)  & 12.38(2.06$\times$)  & 72.03  & 90.99  & 196+48     \\
    SSR-L2,1~\cite{LinJLDL19}          &   1.7(2.41$\times$)  &  12.2(2.09$\times$)  & 71.15  & 90.29  & -     \\
    SSR-L2,0~\cite{LinJLDL19}          &   1.7(2.41$\times$)  &  12.0(2.13$\times$)  & 71.47  & 90.19  & -     \\
    GAL-1~\cite{LinJYZCYHD19}          &  1.58(2.59$\times$)  & 14.67(1.74$\times$)  & 69.88  & 89.75  & 90+60     \\
    GDP~\cite{LinJLWHZ18}              &  1.57(2.61$\times$)  &                   -  & 70.93  & 90.14  & 90+30+20   \\
    HRank~\cite{LinJWZZTL20}           &  1.55(2.64$\times$)  & 13.77(1.86$\times$)  & 71.98  & 91.01  & -     \\
    \textbf{DNAL}($\lambda_a$=7e-5)    &  1.44(2.84$\times$)  & 10.94(2.34$\times$)  & 72.86  & 91.34  & 30+10+70\\
    ABCPruner~\cite{LinJZZWT20}        &  1.30(3.15$\times$)  &                   -  & 72.58  & -      & 12+90\\
    GAL-1-joint~\cite{LinJYZCYHD19}    &  1.11(3.68$\times$)  & 10.21(2.50$\times$)  & 69.31  & 89.12  & 90+60     \\
    ThiNet-30~\cite{LuoZZXWL18}        &  1.10(3.72$\times$)  &  8.66(2.95$\times$)  & 68.17  & 88.86  & 196+48     \\
    HRank~\cite{LinJWZZTL20}           &  0.98(4.17$\times$)  &  8.27(3.09$\times$)  & 69.10  & 89.58  & -     \\
    ABCPruner~\cite{LinJZZWT20}        &  0.94(4.35$\times$)  &                   -  & 70.29  & -      & 12+90\\
    \textbf{DNAL}($\lambda_a$=1e-4)    &  0.88(4.65$\times$)  &  7.18(3.56$\times$)  & 70.17  & 89.78  & 30+10+70\\
    \specialrule{0.10em}{0pt}{0pt}
  \end{tabular}}
\end{table}

\noindent
\textbf{MobileNetV2.}
To further demonstrate  the effectiveness of our DNAL method, we also test it on modern efficient neural networks, i.e., MobileNetV2. The results are presented in Tab.~\ref{tab:6}. Here we compare DNAL with the  state-of-the-art autoML model compression method, i.e. AMC. 
DNAL outperforms AMC by more than 0.2\% with approximate FLOPs (217.24 vs. 211), and even beats it by 0.11\% at smaller computation complexity (207.25 vs. 211). Its ability to compress the lightweight neural networks further is surprising.

\begin{table}[!t]
  \scriptsize
  \caption{Comparison of DNAL  applied to MobileNetV2 with different methods on ImageNet.}
  \label{tab:6}
  \centering
  \setlength{\tabcolsep}{1.8mm}{
  \begin{tabular}{l|rr|rr|r}
    \specialrule{0.10em}{0pt}{0pt}
    Model &
    \makecell*[c]{FLOPs\\ (M)}  &
    \makecell*[c]{Params\\ (M)} &
    \makecell*[c]{Top-1\\ (\%)} &
    \makecell*[c]{Top-5\\ (\%)} &
    \makecell*[c]{Search Cost\\ (Epochs)} \\
    \specialrule{0.10em}{0pt}{0pt}
    Baseline                           &  300.79(1.00$\times$)  & 3.50(1.00$\times$)  & 71.52  & 90.15  & 120     \\
    \specialrule{0.08em}{0pt}{0pt}
    \textbf{DNAL}($\lambda_a$=6e-5)    &  217.24(1.38$\times$)  &  2.87(1.22$\times$) & 71.02  & 89.96  & 80+10+90\\
    AMC~\cite{HeLLWLH18}               &  211(1.43$\times$)     &  -                  & 70.8   & -      & -     \\
    \textbf{DNAL}($\lambda_a$=7e-5)    &  207.25(1.45$\times$)  &  2.78(1.26$\times$) & 70.91  & 89.79  & 80+10+90\\
    \specialrule{0.10em}{0pt}{0pt}
  \end{tabular}}
\end{table}

\noindent
\textbf{ProxylessNAS.}
We show the compression performance of different NAS methods in Tab.~\ref{tab:7}. These searched models are divided into three categories according to the underlying NAS methods used. Our DNAL achieves 75.0\% top-1 and 92.5\% top-5 accuracy on ImageNet with only 3.6M parameters, which is a new state-of-the-art accuracy among different NAS methods. Compared with the EA-based NAS methods, DNAL is more than 2\% and about 1\% higher than CARS in top-1 and top-5 accuracy, respectively, with slightly fewer parameters. It even approximate to the CARS-I model in accuracy with about 1.5$\times$ fewer parameters. Compared to the based-RL NAS methods, DNAL model attains a significantly better performance than both NASNet and MnasNet, while requiring fewer parameters. Our DNAL also surpasses DARTS by 1.7\% and 1.2\% in top-1 and top-5 accuracy, respectively, with significantly fewer parameters, and achieves almost the same top-1 accuracy as ProxylessNAS, but with 2$\times$ fewer parameters. And it achieves higher accuracy and fewer parameters than FBNet. Therefore, it fully demonstrates that our DNAL owns a stronger representability than those state-of-the-art NAS methods.

\begin{table}[!t]
  \scriptsize
  \caption{Comparison of DNAL applied to ProxylessNAS with different methods on ImageNet. Here, EA and RL are the abbreviation for evolutionary algorithm and reinforcement learning, respectively.}
  \label{tab:7}
  \centering
  \setlength{\tabcolsep}{1.8mm}{
  \begin{tabular}{l|r|rr|r|r}
    \specialrule{0.10em}{0pt}{0pt}
    Model &
    \makecell*[c]{Params\\ (M)} &
    \makecell*[c]{Top-1\\ (\%)} &
    \makecell*[c]{Top-5\\ (\%)} &
    \makecell*[c]{Search \\ method} &
    \makecell*[c]{Search Cost\\ (Epochs)} \\
    \specialrule{0.10em}{0pt}{0pt}
    Baseline                            &  16.0   & 75.7   & 92.5   & -         & 150    \\
    \specialrule{0.08em}{0pt}{0pt}
    \textbf{DNAL}($\lambda_a$=6e-5)     &   3.6   & 75.0   & 92.5   & gradient  & 100+10+110\\
    \specialrule{0.08em}{0pt}{0.5pt}
    \specialrule{0.08em}{0.5pt}{0pt}
    CARS-I~\cite{YangWCSXXT19}          &   5.1   & 75.2   & 92.5   & EA        & -     \\
    CARS-A~\cite{YangWCSXXT19}          &   3.7   & 72.8   & 91.6   & EA        & -     \\
    \specialrule{0.08em}{0pt}{0pt}
    NASNet-A~\cite{ZophVSL17}           &   5.3   & 74.0   & 91.6   & RL        & -     \\
    NASNet-B~\cite{ZophVSL17}           &   5.3   & 72.8   & 91.3   & RL        & -     \\
    NASNet-C~\cite{ZophVSL17}           &   4.9   & 72.5   & 91.0   & RL        & -     \\
    MnasNet-92~\cite{TanCPVL18}         &   4.4   & 74.8   & -      & RL        & -     \\
    MnasNet-65~\cite{TanCPVL18}         &   3.6   & 73.0   & -      & RL        & -     \\
    \specialrule{0.08em}{0pt}{0pt}
    DARTS~\cite{LiuSY18}                &   4.7   & 73.3   & 91.3   & gradient  & 600\footnotemark[1]+250     \\
    ProxylessNAS~\cite{CaiZH18}         &   7.1   & 75.1   & 92.5   & gradient  & 200+150     \\
    FBNet-A~\cite{WuDZWSWTVJ19}         &   4.3   & 73.0   & -      & gradient  & 90+360     \\
    FBNet-B~\cite{WuDZWSWTVJ19}         &   4.5   & 74.1   & -      & gradient  & 90+360     \\
    FBNet-C~\cite{WuDZWSWTVJ19}         &   5.5   & 74.9   & -      & gradient  & 90+360     \\
    \specialrule{0.10em}{0pt}{0pt}
  \end{tabular}}
\end{table}
\footnotetext[1]{the number of searching epochs is related to the proxy task, i.e., searching for convolutional cells on CIFAR-10.}

\noindent
\textbf{Efficiency.}
In this part, we further analyze the efficiency of the proposed DNAL method. To demonstrate its efficiency, we choose the number of training epochs, which is hardware-independent, as a metric of learning efficiency, for fair comparison. The experimental results are reported in Tab.~\ref{tab:4},~\ref{tab:5},~\ref{tab:6} and~\ref{tab:7}. We easily observe that DNAL typically features high efficiency. For both VGG16 and ResNet50, to reach optimality, DNAL requires the minimum number of training iterations, with the exception of ABCPruner. DNAL is more than 2.2$\times$ more efficient than ThiNet. For the other two neural networks, i.e., MobileNetV2 and ProxylessNAS, our gradient-based DNAL method is empirically more efficient than the NAS methods based on reinforcement learning (e.g., AMC) and evolutionary optimisation  algorithm (e.g., CARS), which in any case are much expensive in terms of the search cost due to the evaluation of lots of candidate neural architectures. Among the gradient-based NAS methods, DNAL is still  superior in efficiency. 
The cost of searching for the optimal neural architectures of our DNAL method is particularly low, i.e., 20 epochs for CIFAR-10 and 10 epochs for ImageNet.
We argue that the DNAL's high efficiency derives from its differentiability of the objective function thanks to the scaled sigmoid method.

\section{Ablation Study}\label{sec:Ablation_Study}

In this section, we report the results of an ablation study set up to investigate the impact of different factors on both VGG16 and MobileNetV2 using the classification task of CIFAR-10 as a vehicle.

\begin{figure}[!t]
  \centering
  \subfigure[]{
  \includegraphics[trim=5mm 0mm -5mm 0mm, width=1.5in]{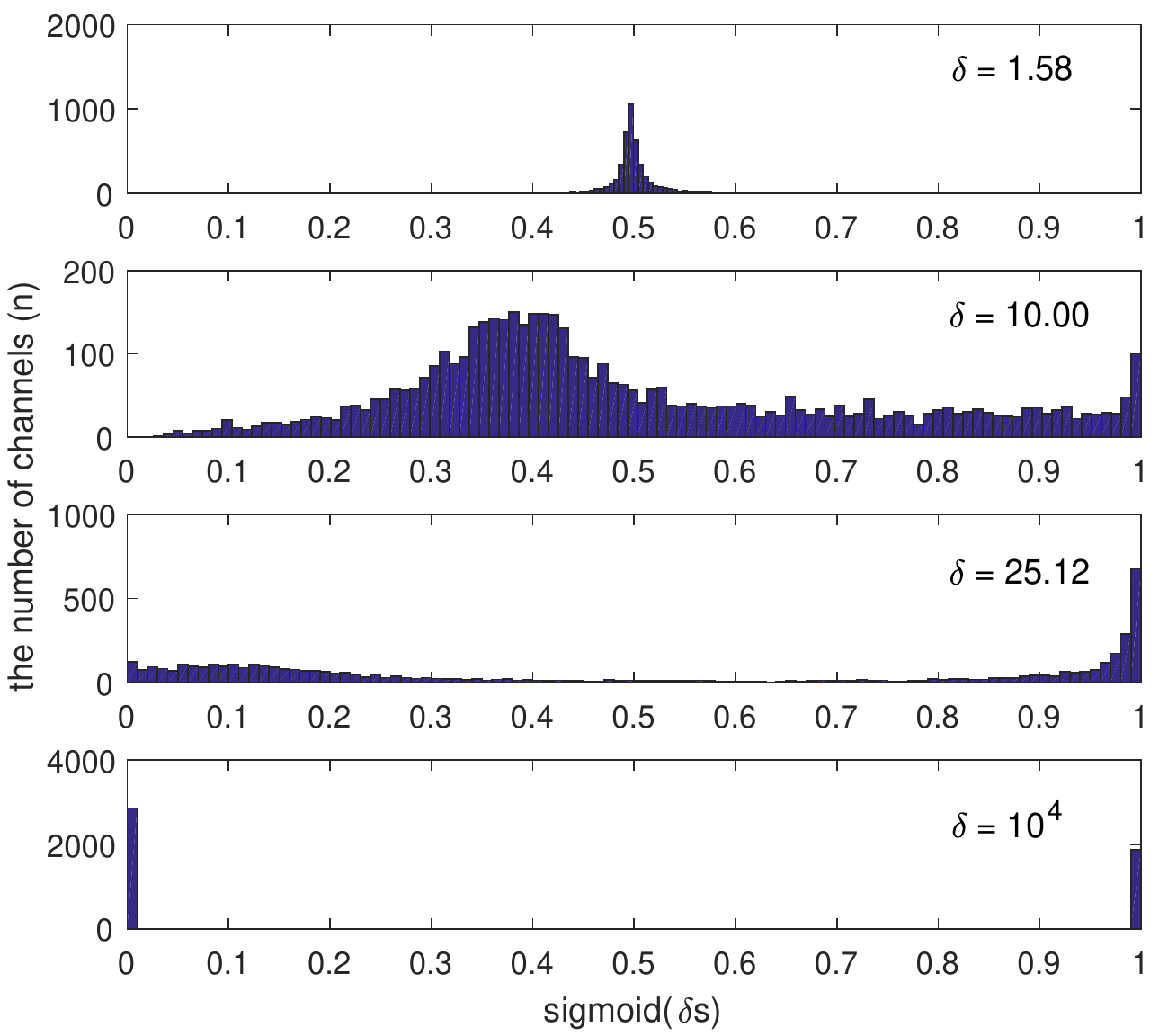}}
  \hspace{0in}
  \subfigure[]{
  \includegraphics[trim=5mm 0mm -5mm 0mm, width=1.5in]{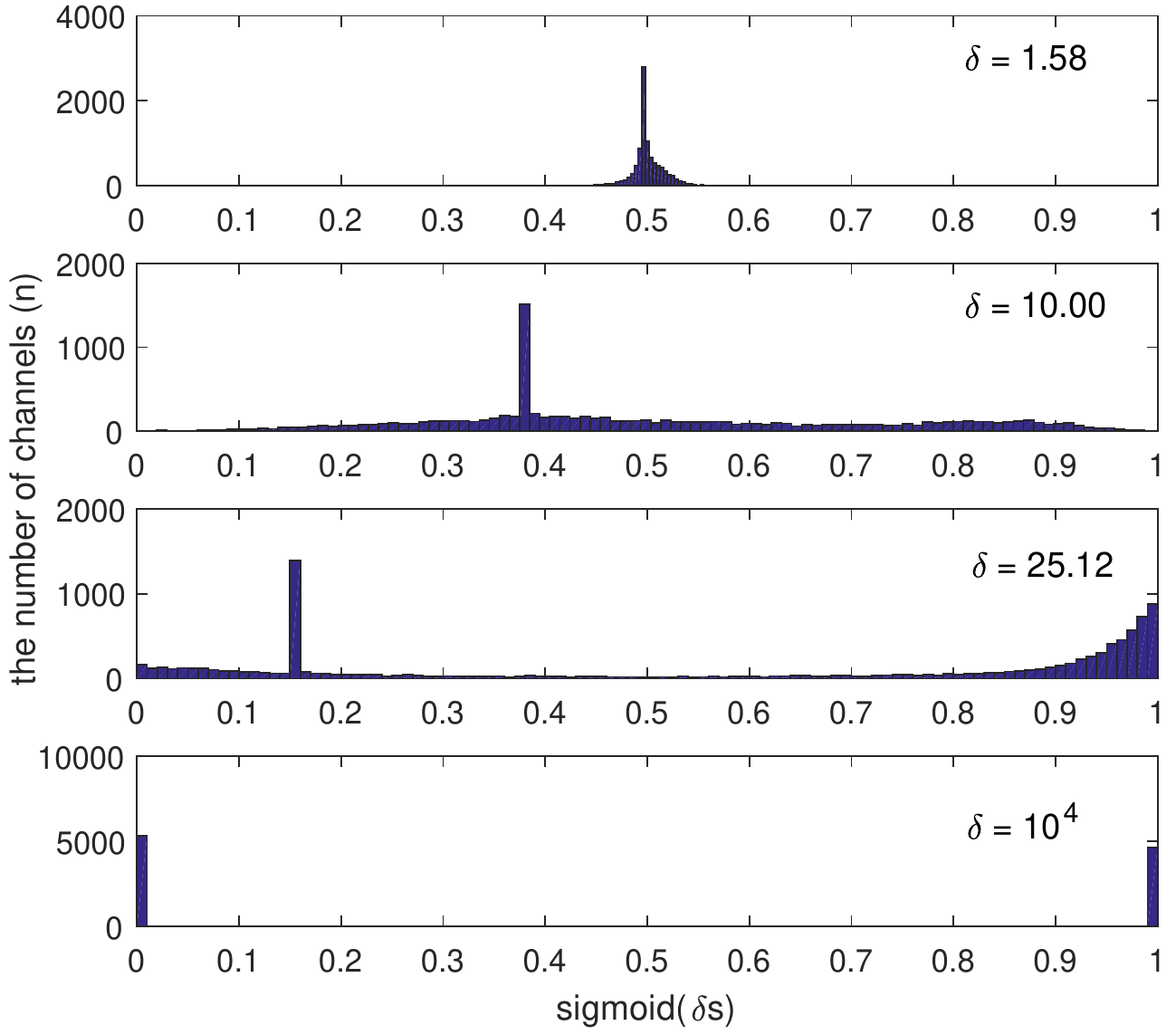}}
  {\caption{Architecture binarization results with different scale factors (\%). (a) VGG16 on CIFAR-10. (b) MobileNetV2 on CIFAR-10.}
  \label{fig:3}}
\end{figure}

\noindent
\textbf{Effect of the scale factor $\delta$.}
As the scale factor increases, the scaled sigmoid function has a larger saturation zone, which helps to binarize the neural architectures. Fig.~\ref{fig:3} shows the distribution of the scaled sigmoid activations obtained with different scale factors.
We initialize the scale factor to $\delta = 1$ and the architecture parameter to $s = 0$ at the beginning of the architecture optimization step, i.e., $sigmoid(\delta s)$ = 0.5.
We can see that the many scaled sigmoid activations are induced to zero under the influence of the scaled sigmoid regularization as the scale factor increases.
When $\delta = 10^4$, all the channels are binarized. The number of channels with $sigmoid(\delta s)$ = 1 defines the efficiency of the resulting neural networks.

\begin{figure}[!t]
  \centering
  \subfigure[]{
  \includegraphics[trim=5mm 0mm -5mm 0mm, width=1.5in]{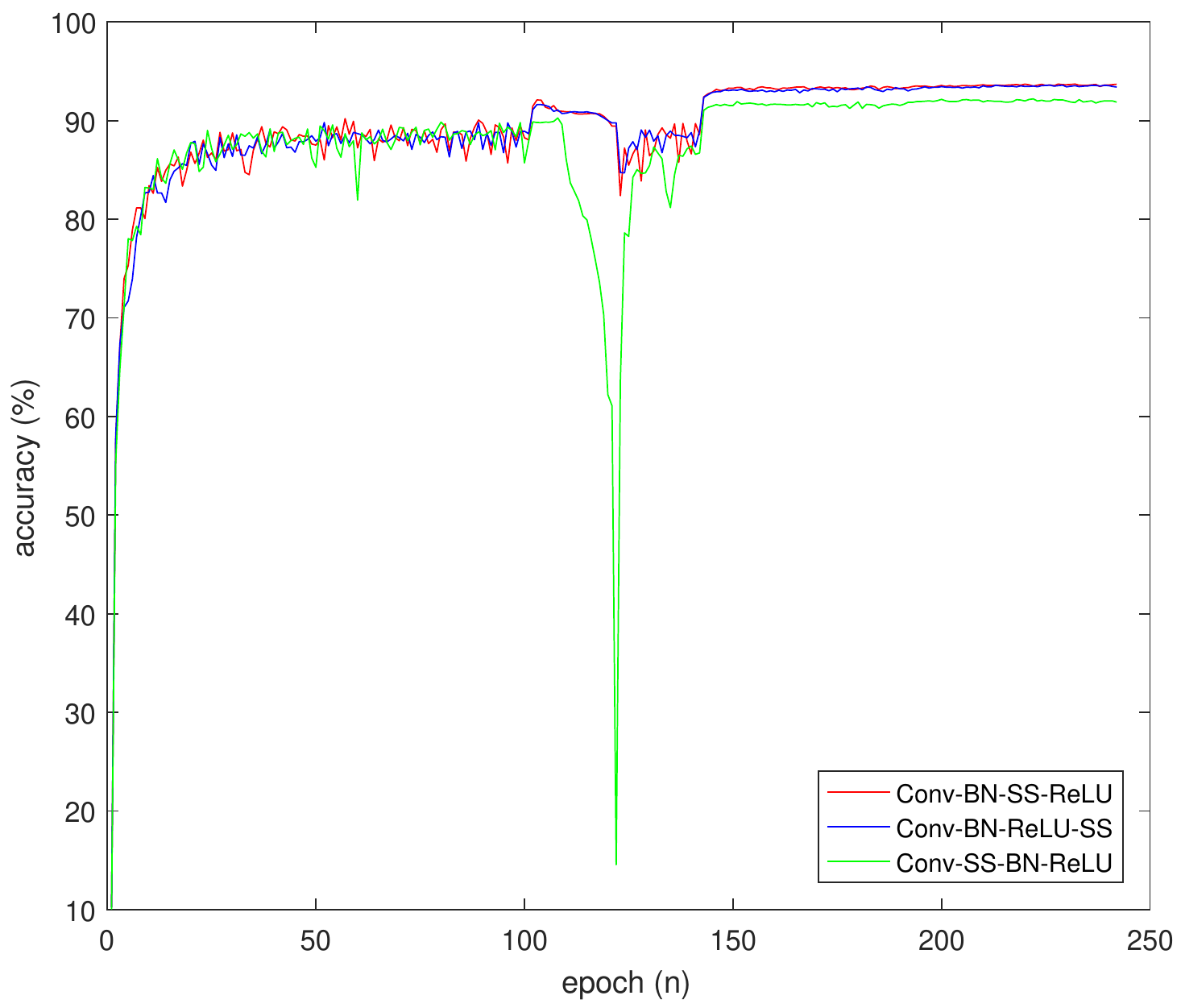}}
  \hspace{0in}
  \subfigure[]{
  \includegraphics[trim=5mm 0mm -5mm 0mm, width=1.5in]{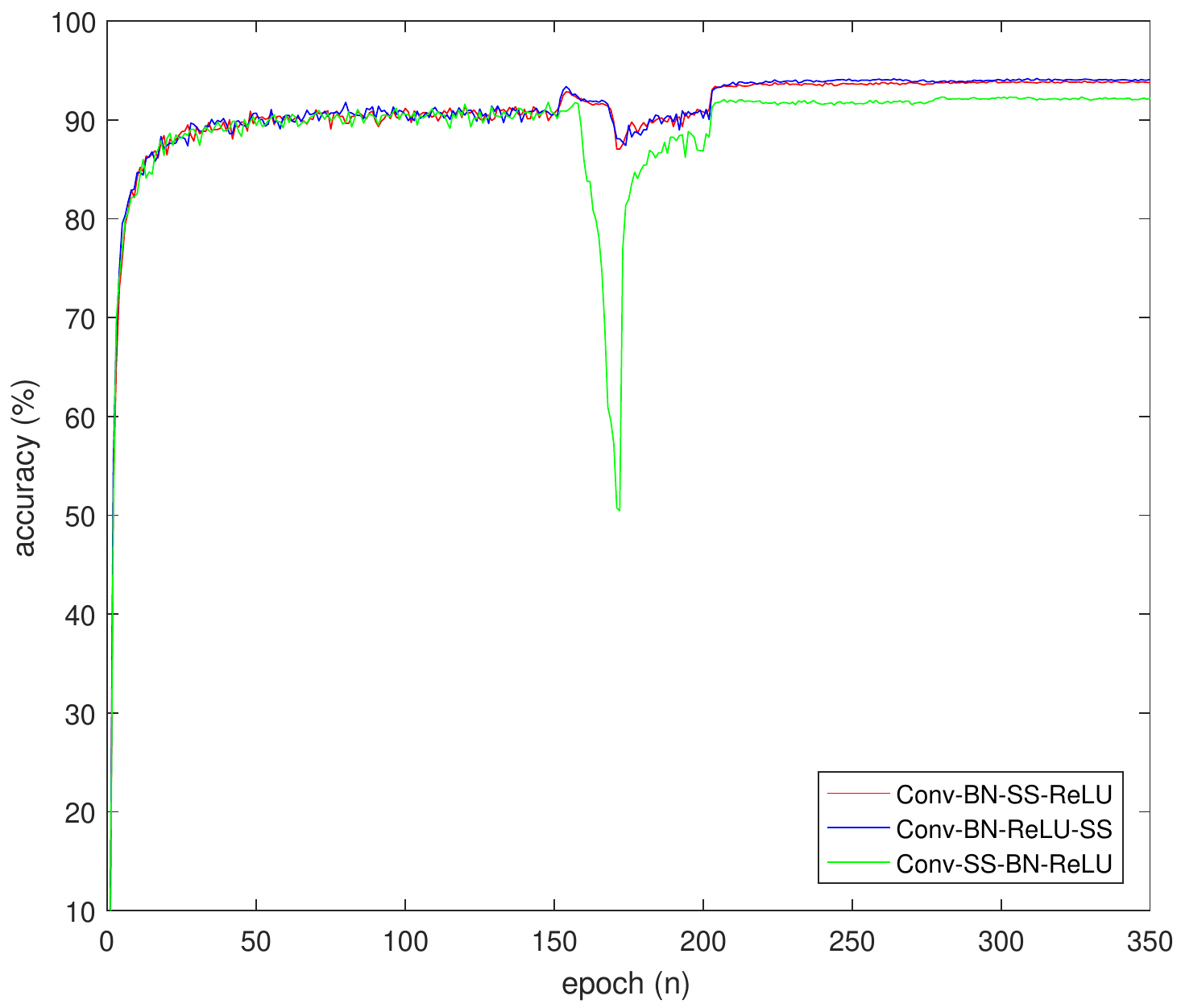}}
  {\caption{Classification accuracy with different configurations (\%). (a) VGG16 on CIFAR-10. (b) MobileNetV2 on CIFAR-10.}
  \label{fig:4}}
\end{figure}

\noindent
\textbf{Effect of the order of $\mathbf{SS}$ layer.}
We explore the effect of different placement of the $\mathbf{SS}$ layer in conjunction with three network configurations, i.e., Conv-SS-BN-ReLU, Conv-BN-SS-ReLU and Conv-BN-ReLU-SS configurations.
Fig.~\ref{fig:4} shows the test accuracy achieved with different network configurations. We observe that both the Conv-BN-SS-ReLU and Conv-BN-ReLU-SS configurations are close in the recognition performance, which is significantly better than Conv-SS-BN-ReLU.
In the weight optimization stage, these three networks are identical because the $\mathbf{SS}$ layer is disabled. Thus, their behaviours are also consistent. In the architecture optimization stage, their network configurations become different due to the enabling $\mathbf{SS}$ layers.
For both the Conv-BN-SS-ReLU and Conv-BN-ReLU-SS configurations, the optimized architectures improve the accuracy at the beginning of the architecture optimization. However, as the number of pruned channels increases, their performance gradually degrades. By contrast, the Conv-SS-BN-ReLU's performance reduces dramatically.
After finetuning, all exhibit improved performance, but both the Conv-BN-SS-ReLU and Conv-BN-ReLU-SS configurations are significantly better than Conv-SS-BN-ReLU.

As we can see, setting the $\mathbf{SS}$ layer behind the BN layer helps to improve the recognition accuracy.
We argue that the BN layer normalizes the distribution of feature maps for each layer, which enables the $\mathbf{SS}$ layer to operate under the same distribution of feature maps.

\begin{figure}[!t]
  \centering
  \subfigure[]{
  \includegraphics[trim=5mm 0mm -5mm 0mm, width=1.5in]{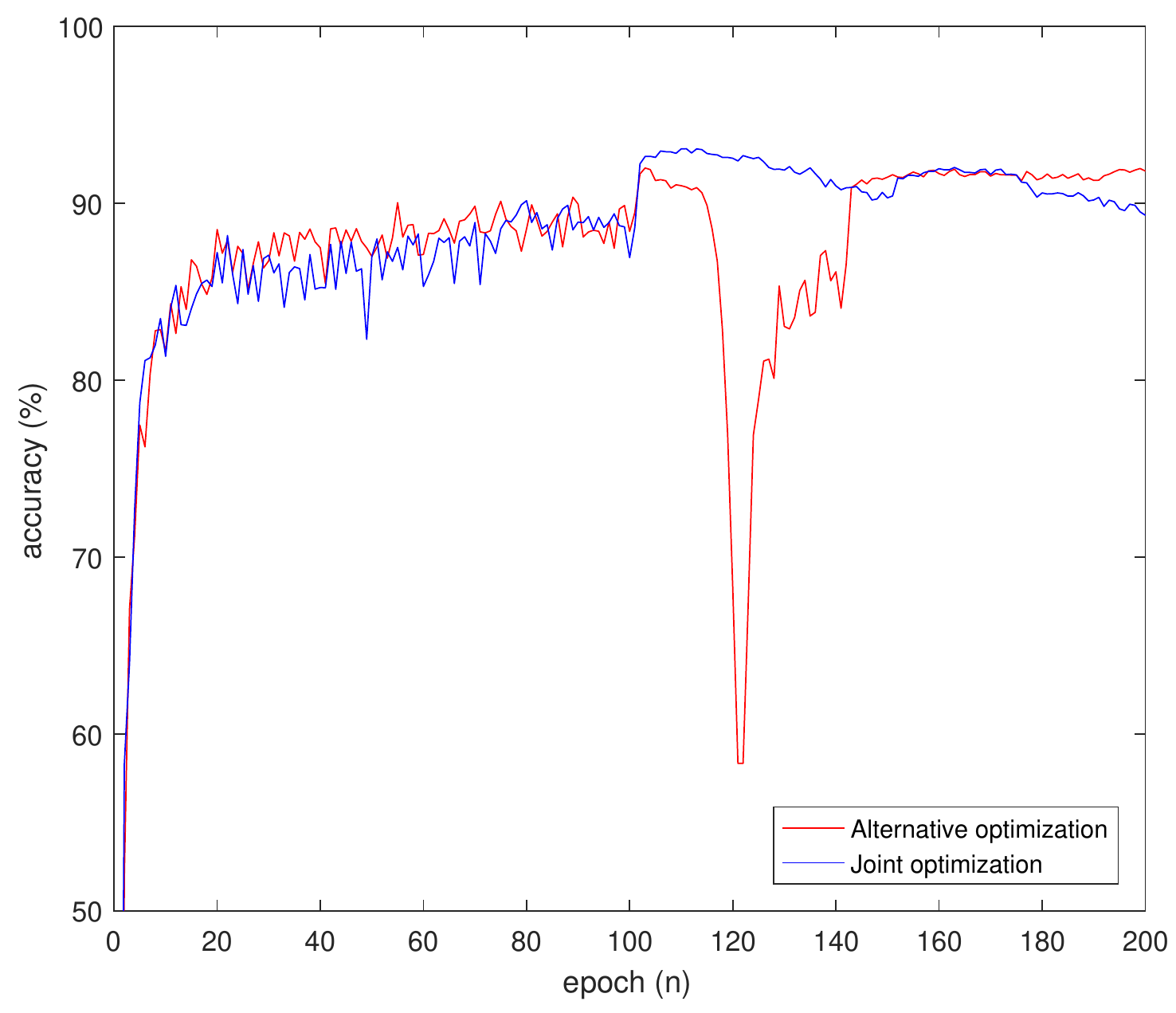}}
  \hspace{0in}
  \subfigure[]{
  \includegraphics[trim=5mm 0mm -5mm 0mm, width=1.5in]{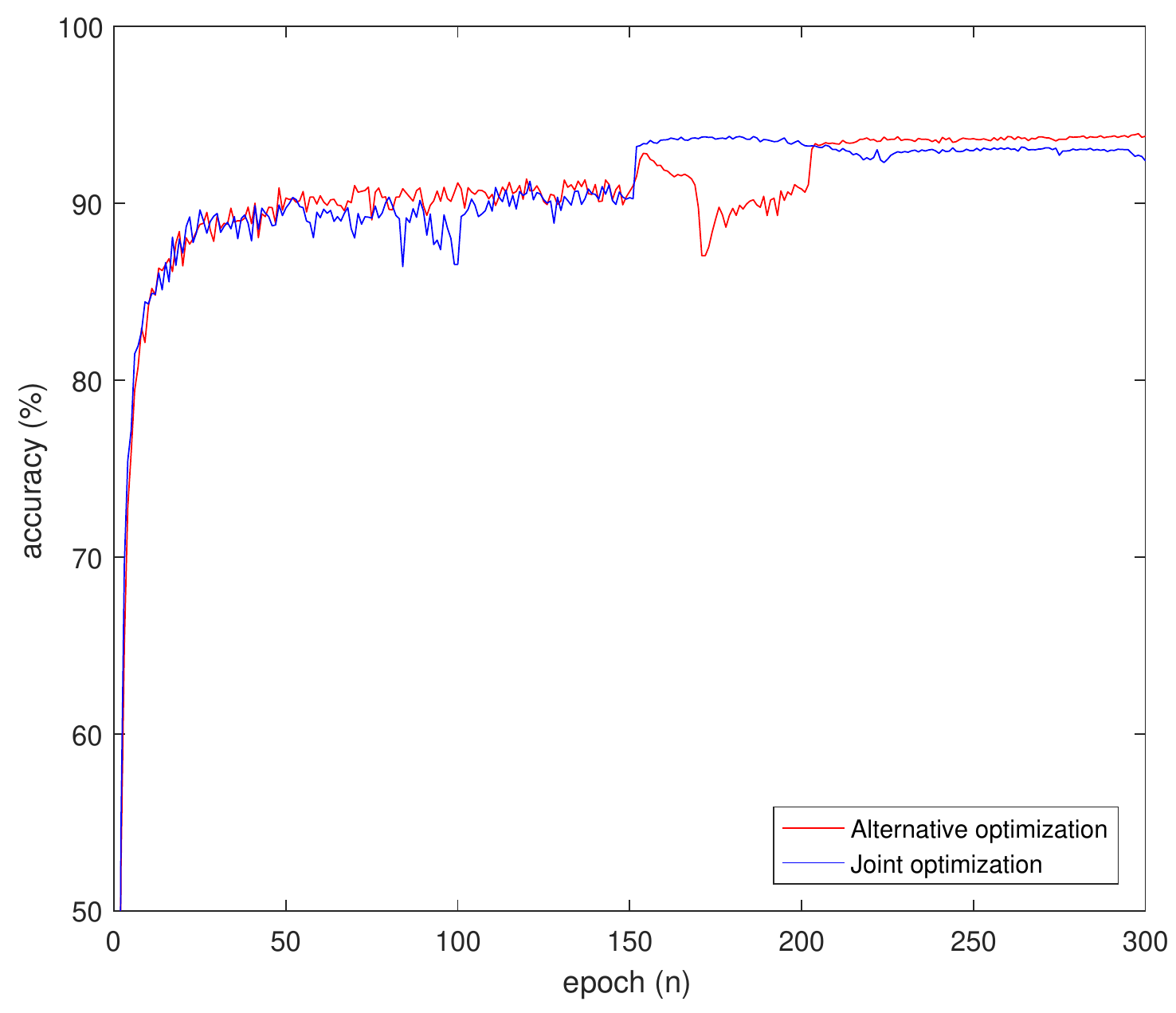}}
  {\caption{Classification accuracy with different optimization strategies but with but similar efficiency (\%). (a) VGG16 on CIFAR-10. (b) MobileNetV2 on CIFAR-10.}
  \label{fig:5}}
\end{figure}

\noindent
\textbf{Effect of the joint optimization.}
In this paper, we optimize the weights and architecture parameters in a sequential manner. However, they can be optimized jointly.
We compare these two optimization strategies in Fig.~\ref{fig:5}. For a fair comparison, the searched models are similar in the computational efficiency.
At the beginning of the network optimization, the joint strategy has a faster convergence, and exhibits better accuracy. However, as the network continuous to be optimised, the performance gradually degrades in accuracy. For the sequential strategy, after the architecture optimization, the resulting network rapidly recovers its performance by finetuning, surpassing the joint strategy optimization in accuracy.

This confirms that the joint optimization is hindered by the vanishing gradient problem. With the increasing scale factors $\delta$, more channels enter the saturation zone where the gradients of the architecture parameters are zero. This results in inefficient network optimization and in consequence a loss of accuracy. After the architecture optimization, DNAL disables the $\mathbf{SS}$ layer and optimizes only the weights, which avoids the vanishing gradient problem caused.

\section{Conclusion}\label{sec:Conclusion}
We have presented a differentiable neural architecture learning method (DNAL). DNAL utilizes the scaled sigmoid function to relax the discrete architecture space into a continuous architecture space, and gradually converts the continuous optimization problem into the binary optimization problem. The optimal neural architecture is learned by gradient-based methods without the need to evaluation candidate architectures individually, thus significantly improving the search efficiency.
We introduced a new $\mathbf{SS}$ module layer to implement the scaled sigmoid activation function, enriching the module family of neural networks for the optimization of neural architectures. The proposed DNAL method was applied to conventional CNNs, lightweight CNNs and stochastic supernets.
Extensive experiments on CIFAR-10 and ImageNet-1K demonstrated that DNAL delivers state-of-the-art performance in terms of accuracy, model size and computational complexity, especially search cost.

\section*{Acknowledgment}
This work is supported by the National Key R\&D Program of China (Grant No. 2018YFB1004901), by the National Natural Science Foundation of China (Grant No.61672265, U1836218), by the 111 Project of Ministry of Education of China (Grant No. B12018), by UK EPSRC GRANT EP/N007743/1, MURI/EPSRC/DSTL GRANT EP/R018456/1.



\bibliographystyle{IEEEtran}
\bibliography{mydeeplib}


\begin{IEEEbiography}[{\includegraphics[width=1in,height=1.25in,clip,keepaspectratio]{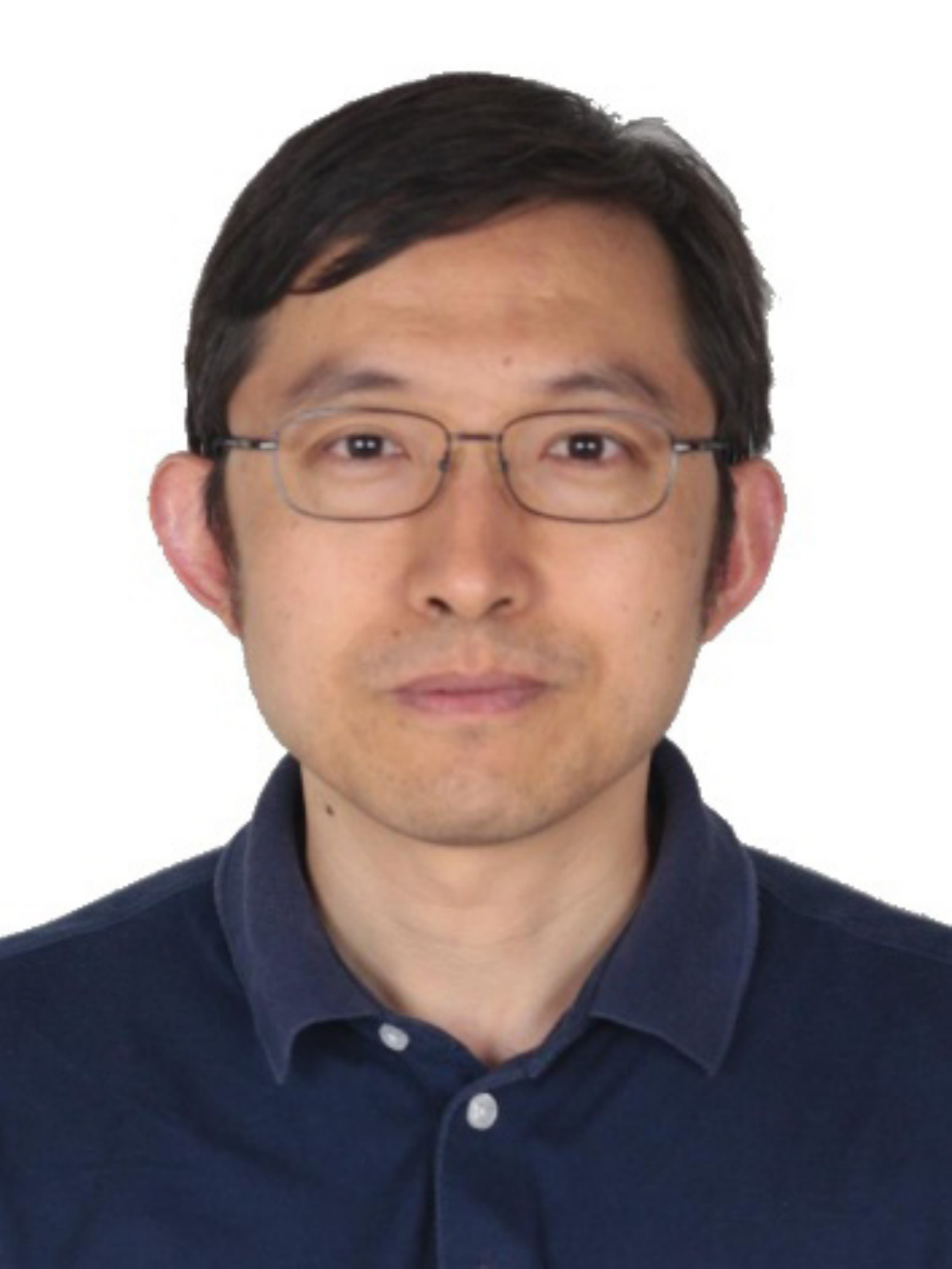}}]{Qingbei Guo}
received the M.S. degree from the School of Computer Science and Technology, Shandong University, Jinan, China, in 2006. He is a member of the Shandong Provincial Key Laboratory of Network based Intelligent Computing and the lecturer in the School of Information Science and Engineering, University of Jinan. He is now a Ph.D. student at Jiangnan University, Wuxi, China. His current research interests include wireless sensor networks, deep learning/machine learning, computer vision and neuron networks.
\end{IEEEbiography}

\begin{IEEEbiography}[{\includegraphics[width=1in,height=1.25in,clip,keepaspectratio]{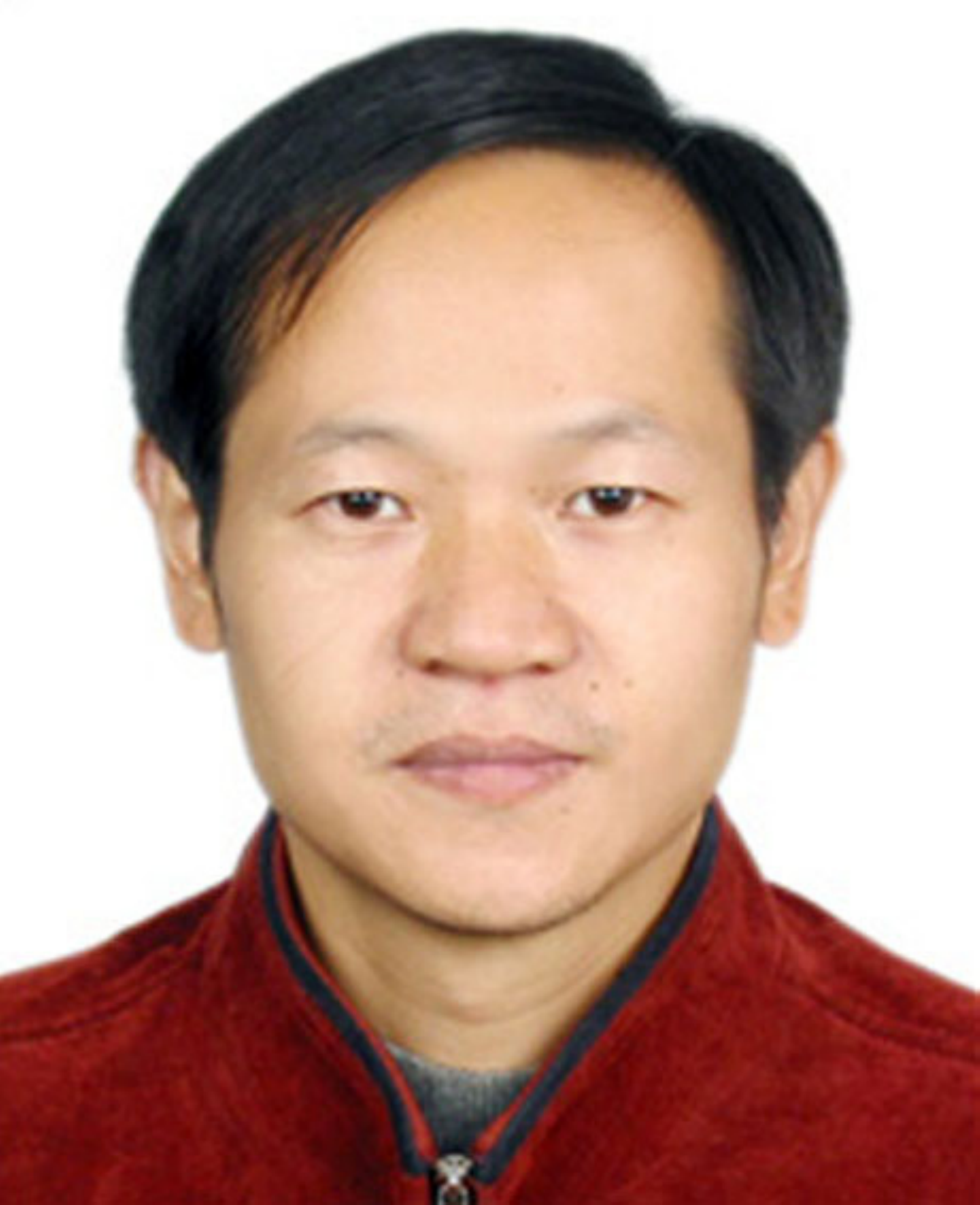}}]{Xiao-jun Wu}
received his B.S. degree in mathematics from Nanjing Normal University, Nanjing, PR China in 1991 and M.S. degree in 1996, and Ph.D. degree in Pattern Recognition and Intelligent System in 2002, both from Nanjing University of Science and Technology, Nanjing, PR China, respectively. He was a fellow of United Nations University, International Institute for Software Technology (UNU/IIST) from 1999 to 2000. From 1996 to 2006, he taught in the School of Electronics and Information, Jiangsu University of Science and Technology where he was an exceptionally promoted professor. He joined the School of Information Engineering, Jiangnan University in 2006 where he is a professor. He won the most outstanding postgraduate award by Nanjing University of Science and Technology. He has published more than 300 papers in his fields of research. He was a visiting researcher in the Centre for Vision, Speech, and Signal Processing (CVSSP), University of Surrey, UK from 2003 to 2004. His current research interests are pattern recognition, computer vision, fuzzy systems, neural networks and intelligent systems.
\end{IEEEbiography}

\begin{IEEEbiography}[{\includegraphics[width=1in,height=1.25in,clip,keepaspectratio]{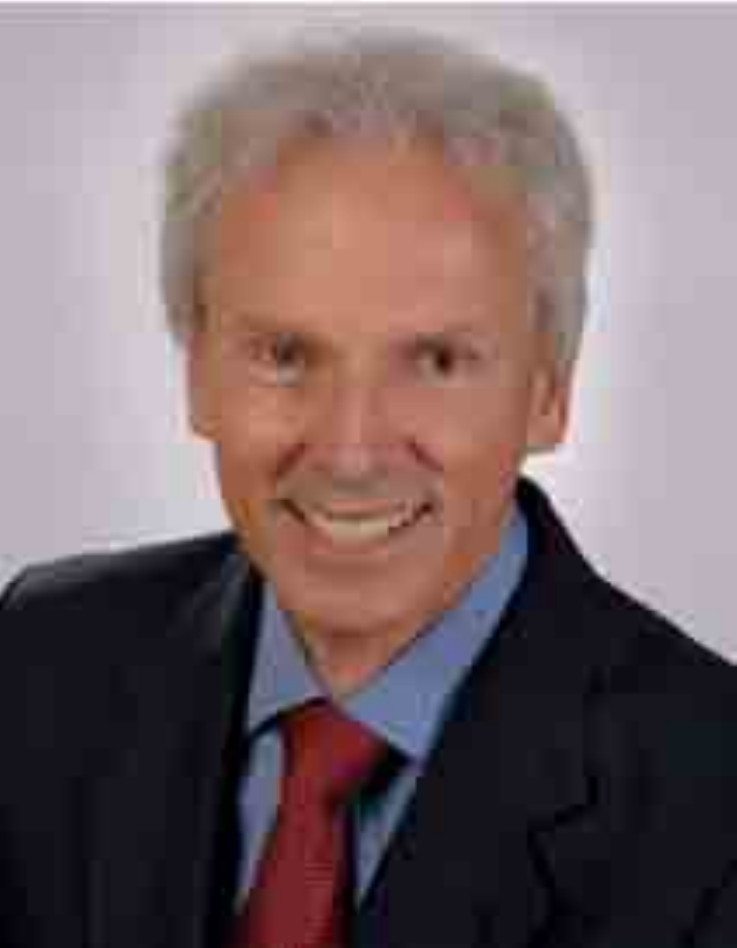}}]{Josef Kittler}
received the B.A., Ph.D., and D.Sc. degrees from the University of Cambridge, in 1971, 1974, and 1991, respectively. He is a distinguished Professor of Machine Intelligence at the Centre for Vision, Speech and Signal Processing, University of Surrey, Guildford, U.K. He conducts research in biometrics, video and image database retrieval, medical image analysis, and cognitive vision. He published the textbook Pattern Recognition: A Statistical Approach and over 700 scientific papers. His publications have been cited more than 66,000 times (Google Scholar). He is series editor of Springer Lecture Notes on Computer Science. He currently serves on the Editorial Boards of Pattern Recognition Letters, Pattern Recognition and Artificial Intelligence, Pattern Analysis and Applications. He also served as a member of the Editorial Board of IEEE Transactions on Pattern Analysis and Machine Intelligence during 1982-1985. He served on the Governing Board of the International Association for Pattern Recognition (IAPR) as one of the two British representatives during the period 1982-2005, President of the IAPR during 1994-1996.
\end{IEEEbiography}

\begin{IEEEbiography}[{\includegraphics[width=1in,height=1.25in,clip,keepaspectratio]{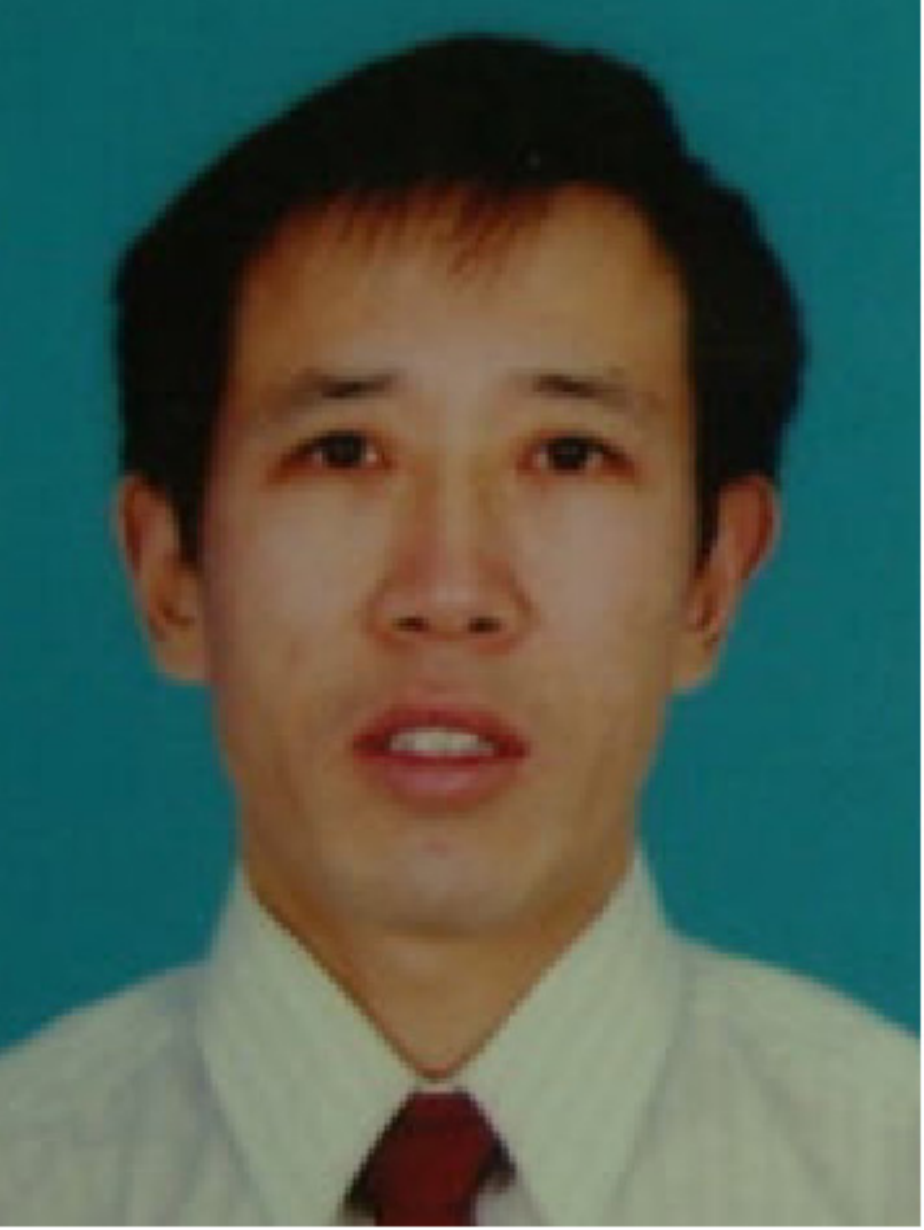}}]{Zhiquan Feng}
received his M.S. degree in Computer Software from Northwestern Polytechnical University, China, in 1995, and Ph.D. degree in Computer Science \& Engineering from Shandong University, China, in 2006. He is currently a Professor at University of Jinan, China. Dr. Feng is a visiting Professor of Sichuang Mianyang Normal University. As the first author or corresponding author he has published more than 100 papers in international journals and conference proceedings, 2 books, and 30 patents in the areas of human hand recognition and human-computer interaction. He has served as the Deputy Director of Shandong Provincial Key Laboratory of network based Intelligent Computing, group leader of Human Computer Interaction based on natural hand, editorial board member of Computer Aided Drafting Design and Manufacturing, CADDM, and also an editorial board member of The Open Virtual Reality Journal. He is a deputy editor of World Research Journal of Pattern Recognition and a member of Computer Graphics professional committee. Dr. Feng’s research interests are in human hand tracking/recognition/interaction, virtual reality, human-computer interaction, and image processing. His research has been extensively supported by the Key R\&D Projects of the Ministry of Science and Technology, Natural Science Foundation of China, Key Projects of Natural Science Foundation of Shandong Province, and Key R\&D Projects of Shandong Province with total grant funding over three million RMB. For more information, please refer to http://nbic.ujn.edu.cn/nbic/index.php.
\end{IEEEbiography}


\begin{appendices}
\end{appendices}

\end{document}